\documentclass[letterpaper, 10 pt, conference]{ieeeconf}  

\usepackage{amssymb}
\usepackage{xcolor}
\usepackage{graphicx}
\usepackage{subcaption}
\usepackage{multirow}
\newcommand{\ind}{\mathbf{1}}
\newcommand{\update}{\textcolor{black}}
\usepackage{amsmath}
\usepackage{makecell}
\usepackage{hyperref}
\usepackage{diagbox}
\usepackage{booktabs}
\usepackage{boldline}
\usepackage{microtype}  
\AtBeginDocument{
  \setlength{\abovedisplayskip}{4pt plus 1pt minus 2pt}
  \setlength{\belowdisplayskip}{4pt plus 1pt minus 2pt}
  \setlength{\abovedisplayshortskip}{0pt plus 1pt}
  \setlength{\belowdisplayshortskip}{2pt plus 1pt minus 1pt}
}
\IEEEoverridecommandlockouts                              

\title{\LARGE \bf
CADRE: Dynamic Catching via Implicit Contact Descriptors and Task-Appropriate Recovery Affordances
}
\author{
\authorblockN{Fan Yang$^{1}$, Zixuan Huang$^{1}$, Abhinav Kumar$^{1}$,\\ Sergio Aguilera Marinovic$^{2}$, Soshi Iba$^{2}$, Rana Soltani Zarrin$^{2}$, Dmitry Berenson$^{1}$}
\vspace{-1cm}
\thanks{This work was supported by Honda Research Institute USA.}
\thanks{$^{1}$ University of Michigan, Ann Arbor. $^{2}$ Honda Research Institute, USA 
}
}
\begin{document}
\maketitle
\begin{abstract}
Real-world dexterous manipulation often encounters unexpected errors and disturbances, which can lead to catastrophic failures, such as dropping the manipulated object. 
To address this challenge, we focus on the problem of catching a falling object while it remains within grasping range and, importantly, resetting the system to a configuration favorable for resuming the primary manipulation task.
We propose \textbf{Contact-Aware Dynamic Recovery (CADRE)}, a reinforcement learning framework that incorporates a Neural Descriptor Field (NDF)-inspired module to extract implicit contact features. 
Building on these contact features, we introduce an Implicit Recovery Affordance function to encourage recovery to task-appropriate states.
Compared to methods that rely solely on object pose or point cloud input, NDFs can directly reason about finger-object correspondence and better establish a recovery target for RL training. 
Our experiments show that incorporating contact features improves training efficiency, enhances convergence performance for RL training, and ultimately leads to more successful recoveries. 
Additionally, we demonstrate that CADRE can generalize zero-shot to unseen objects with different geometries.
\end{abstract}
\vspace{-0.3cm}
\section{Introduction}
\vspace{-0.2cm}
Robots performing real-world manipulation can encounter unexpected disturbances and modeling errors, leading to catastrophic failure, such as dropping the 
manipulated object. 
While significant progress has been made in improving robustness~\cite{tobin2017domain, kumar2021rma}, real-world applications often present unpredictable disturbances and unmodeled factors that exceed the system's designed robustness tolerance. 
For instance, during a screwdriver-turning task, a stuck screw can generate unexpectedly large torques on the screwdriver, causing the fingers to slip and the object to fall. 
This issue is further exacerbated in robotic systems without tactile sensing, where the absence of direct contact feedback makes it challenging to detect whether the object is being grasped firmly.
\begin{figure}[htbp]
    \centering
    \includegraphics[width=1.0\linewidth]{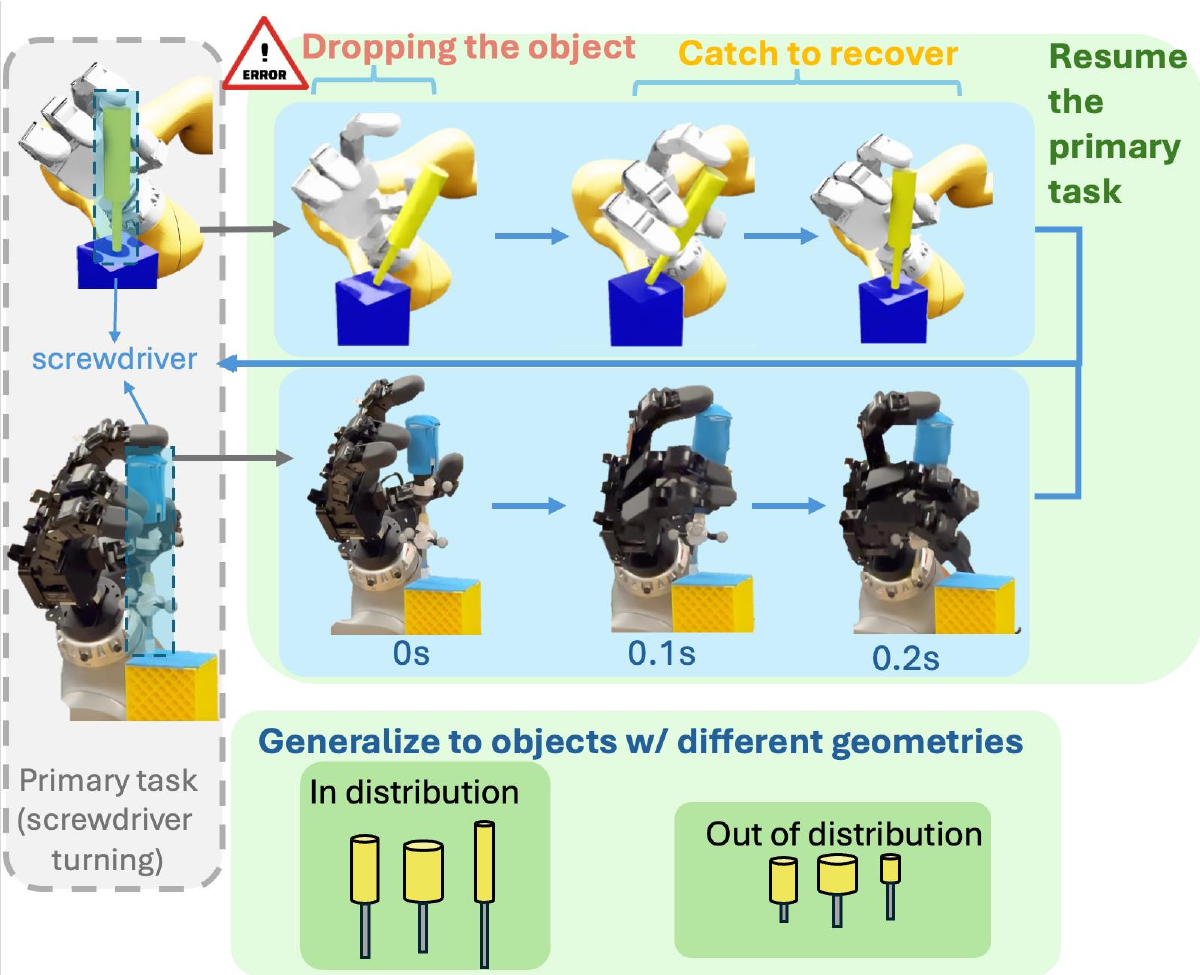}
    \caption{CADRE recovers from object-dropping failures by catching the object and resetting to states favorable for resuming the primary manipulation task. Leveraging contact information from a pre-trained NDF model, it generalizes recovery behaviors to objects with different geometries.}
    \label{fig:teaserl}
    \vspace{-0.7cm}
\end{figure}
In this work, we focus on dynamic recovery, specifically, catching falling objects before irrecoverable failure occurs. Rather than improving the inherent robustness of primary manipulation policies, we propose a complementary strategy that incorporates a fallback catching policy; i.e. the robot switches from the primary manipulation policy to the catching policy when object dropping is detected. This policy is responsible for catching falling objects during failure events, and, importantly, resetting the robot system to states from which the primary manipulation can resume. This is because merely catching falling objects is not enough to ensure successful recovery. Specifically, while power grasps can effectively catch a large variety of objects~\cite{kim2014catching, feix2015grasp}, the primary manipulation task often requires specific grasp types, such as precision grasps~\cite{tang2024robotic, zarrin2023hybrid}. Switching from power grasps to precision grasps presents significant challenges. Therefore, our recovery policy is designed to achieve grasp configurations that support the contact requirements of the primary manipulation task. Additionally, to ensure practical applicability, the recovery policy must be able to adapt to objects with different geometries.

To address the challenges of dynamic recovery and generalization across shapes, we present Contact-Aware Dynamic Recovery (CADRE), a reinforcement learning approach that incorporates contact information in its observation space and reward definition. 
Our work is based on the importance of contact in dexterous manipulation~\cite{jin2024complementarity, cheng2021contact, grady2021contactopt}. 
CADRE is motivated by the insight that maintaining consistent contact behaviors across different object geometries is one of the fundamental factors for successful generalization. To achieve this capability, contact information is derived from Neural Descriptor Fields (NDF)~\cite{simeonov2022neural}, which captures the geometric correspondence between 3D coordinates and the object point clouds. CADRE leverages NDF features as implicit contact information for dexterous manipulation. NDF features of a predefined set of keypoints on the hand are used to characterize the grasp configuration. This approach provides comprehensive contact modeling for both regions that should be in contact (e.g., fingertips) and regions where contact should be avoided (e.g., palm). Crucially, CADRE uses this same implicit representation not only to observe the grasp, but to formulate the recovery objective. To define task-appropriate recovery states, we propose to learn an Implicit Recovery Affordance (IRA) from a small set of task demonstrations. By applying Noise Contrastive Estimation (NCE)~\cite{gutmann2010noise} to these NDF-encoded demonstrations, we contrastively learn a dense energy landscape of successful grasps for recovery. This learned affordance guides the policy to catch objects using task-appropriate grasps despite changes in object geometries. 
\looseness=-1

Our main contributions are summarized as follows: (1) We propose the problem of recovery through catching, where a robot must not only catch the falling object but also achieve grasp configurations from which the robot can seamlessly resume the primary manipulation task; (2) We develop an NDF-based implicit contact representation for contact-rich dexterous manipulation that effectively captures the geometric correspondences between the hand and the manipulated object; (3) We introduce an \textbf{Implicit Recovery Affordance (IRA)} within our RL framework, leveraging NCE to learn a dense energy landscape of desired grasps from limited demonstrations to serve as a reward signal for task-appropriate recovery; (4) We demonstrate empirically that our contact representation enables effective generalization across different geometries in dynamic recovery tasks.

Our experimental results demonstrate that our contact-aware approach significantly improves recovery performance on training objects while enabling zero-shot generalization to unseen objects of the same type but with different geometries (e.g. various sizes of screwdriver). Please see more details and \textbf{appendix} at \href{https://cadrecatching.github.io/}{https://cadrecatching.github.io/}.
\section{Related Work}
\subsection{Dynamic Manipulation}
Dynamic manipulation involving rapid robot and object motion has been a popular research area~\cite{ishihara2006dynamic, zeng2020tossingbot, yang2024dynamic, hou2020robust}. Catching fast-moving objects is a particularly relevant subdomain to our work. Prior work has explored both planning-based approaches~\cite{wang2024caging, namiki2003development, kim2014catching, salehian2016dynamical} and RL-based methods~\cite{zhang2024catch, huang2023dynamic, lan2023dexcatch, charlesworth2021solving} for object catching. However, these methods primarily focus on stable catching but without considering grasp configurations, often converging to power grasps and more importantly, overlooking subsequent manipulation task requirements. In contrast, our recovery-through-catching framework addresses a more difficult challenge: where the robot must not only catch falling objects but also achieve grasp configurations that enable seamless resumption of the primary manipulation task. 
Additionally, in our recovery task, the object spends considerably less time in the air, demanding faster and more dynamic arm motions.
\subsection{Representation Learning for Manipulation}
Perception representations (e.g., point cloud, image, contact information) significantly impact RL manipulation performance. 
A majority of research directly uses point clouds as the representation for the 3D scene~\cite{qin2023dexpoint, huang2021generalization, wu2023learning, bao2023dexart}. Recent work has also explored more sophisticated representations. For instance, Wu et al.~\cite{wu2023daydreamer} encode image observations into a learned latent space for model-based RL, while Driess et al.~\cite{driess2022reinforcement} utilize Neural Radiance Fields~\cite{mildenhall2021nerf} to obtain latent scene embeddings for RL policy inputs. However, those methods do not consider contact-rich dexterous manipulation scenarios. 
Yang and Jin~\cite{yang2025contactsdf} use signed distance functions to model contacts but do not identify task-appropriate geometries. Wei et al.~\cite{wei2024mathcal} and Wang et al.~\cite{wang2024neural} propose highly related grasp generation methods based on learned representations derived from robot and object point clouds. However, these approaches target quasi-static grasps, unlike our highly dynamic catch-for-recovery focus.

Implicit geometric representations, like Neural Descriptor Fields (NDFs), show promise in manipulation for cross-embodiment generalization~\cite{khargonkar2023neuralgrasps} and motion planning~\cite{cheng2023nod, huang2024implicit, simeonov2023se}. However, the application of such implicit geometric representations to RL and dynamic dexterous manipulation remains unexplored. Our work addresses this gap by leveraging NDF-inspired contact representations to enable dynamic recovery. 
\looseness=-1
\vspace{-0.1cm}
\section{Problem Statement}
\vspace{-0.1cm}
\label{sec:problem_statement}
We focus on the task of catching a falling object for recovery. Specifically, the recovery problem inherently includes two objectives: (1) the robot must prevent dropping the object, and (2) the system should recover to a state from which the primary manipulation task can be resumed. The inclusion of the second objective distinguishes our work from prior literature on catching. 

We formulate the dynamic recovery as a Markov Decision Process (MDP). 
At each time step, the policy receives observations $\mathbf{o}_t:= \{\mathbf{q}_t, \mathbf{x}_t, \mathbf{v}_t \}$, where $\mathbf{q}_t$ represents the robot's joint angles, $\mathbf{x}_t$ is the object pose in $\mathrm{SE}(3)$, and $\mathbf{v}_t$ denotes the object's twist (linear and angular velocities). We assume access to a low-level joint position controller; therefore, the robot action $\mathbf{a}_t$ is defined as the desired joint position at the next time step. 

To make the problem tractable, we assume knowledge of the object's geometry, represented by a point cloud $\mathbf{P}$, which can be generated from a scan or a CAD model of the object. This assumption is reasonable in a factory setting, where the set of available objects, such as screwdrivers, is usually fixed and the geometry of each object can be obtained.
The objective is to recover the system to a task-appropriate configuration. Rather than predefining a target pose, we formulate the recovery goal as a manifold of valid grasps derived from a small set of primary manipulation demonstrations collected on a single object.

Additionally, we aim to develop a method that generalizes across object geometries. During RL training, objects are randomly sampled from a predefined distribution. While during evaluation, objects are selected from both the in-distribution (ID) set and the out-of-distribution (OOD) set. An ideal method should be able to catch the object in contact configurations similar to those from the training examples, even if the target object is out of distribution. 

The method is evaluated based on (1) whether the robot successfully catches the object, (2) and the performance of the subsequent primary manipulation task. Compared to other dexterous manipulation setups, this task is more challenging because the robot needs to move quickly to handle dynamic situations, but also precisely to reach the desired configurations. 
\looseness=-1
\vspace{-0.3cm}
\section{Methods}
\vspace{-0.1cm}
\begin{figure*}
\vspace{0.2cm}
    \centering
    \includegraphics[width=0.8\linewidth]{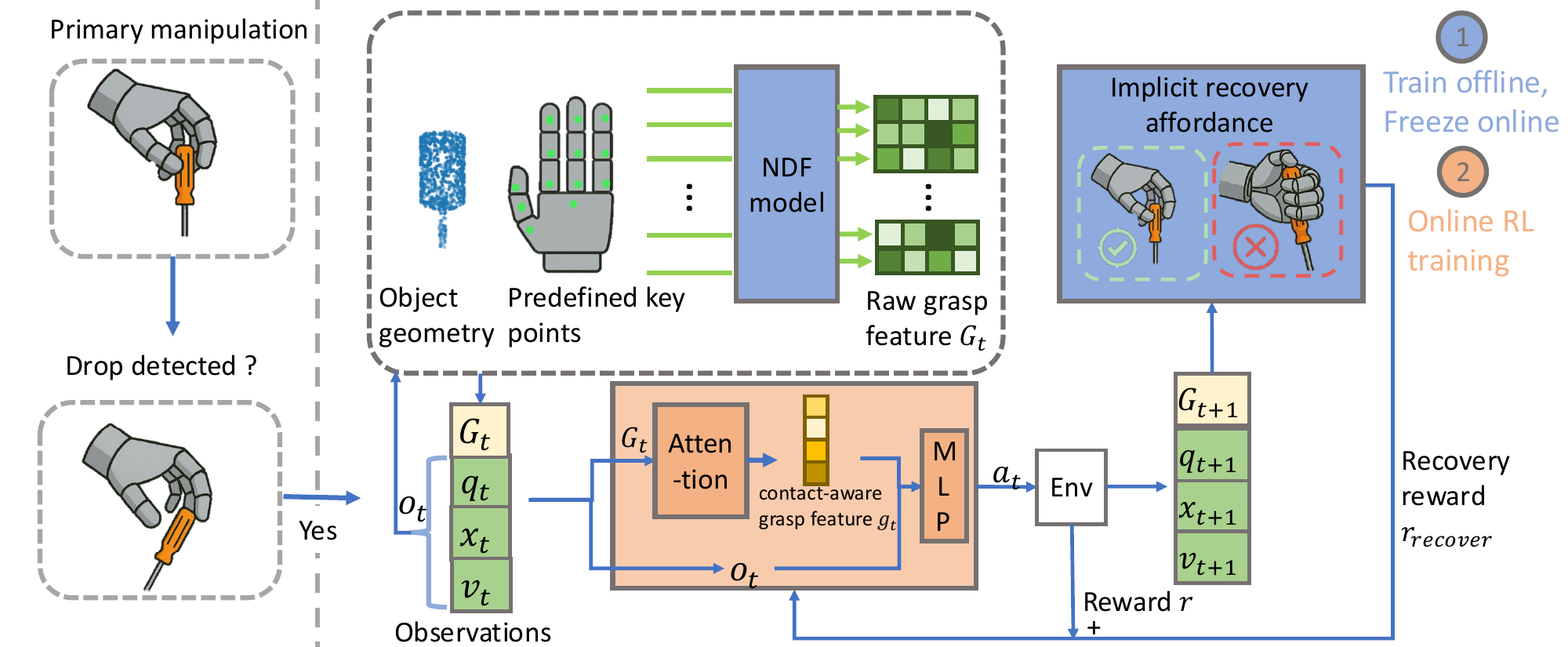}
    \vspace{-0.1cm}
    \caption{Contact-Aware Dynamic Recovery (CADRE) aims to catch a manipulated object when it falls from the grasp and recover to states that support the resumption of the manipulation task. CADRE leverages a pretrained NDF model to extract implicit contact features from the grasp. The contact-aware grasp features are incorporated into the RL observations, enhancing the policy's awareness of contact and improving recovery performance.}
    \label{fig:method}
    \vspace{-0.7cm}
\end{figure*}
We propose Contact-Aware Dynamic Recovery (CADRE), which leverages reinforcement learning to optimize recovery policies in dexterous manipulation tasks. 
To enable contact awareness and generalization to unseen objects based on contacts, we incorporate contact features derived from NDFs into both observation space and reward definition. The NDF model extracting contact features is pretrained and remains fixed during RL training.
The contact features enable the RL policy to achieve consistent behaviors across objects with different geometries. 
\subsection{Preliminary: Neural Descriptor Fields}
Neural Descriptor Fields (NDFs)~\cite{simeonov2022neural} is a learned representation that captures geometric correspondence between a queried 3D coordinate and an object. Given an object point cloud $\mathbf{P}$, NDF extracts an n-dimensional geometric feature for any 3D coordinate $\mathbf{p}$:
\begin{equation}
    f_{NDF}(\mathbf{p}|\mathbf{P}): \mathbb{R}^3 \rightarrow \mathbb{R}^n.
\end{equation}
During NDF training, an MLP-based network $\Phi(\mathbf{p}|\mathcal{E}(\mathbf{P}))$ is trained to predict occupancy or signed distance of the query point $\mathbf{p}$, conditioned on the object point cloud $\mathbf{P}$. $\mathcal{E}$ is the PointNet-based encoder, which extracts the latent features of the object point cloud. The NDF feature is defined as:
\begin{equation}
    f_{NDF}(\mathbf{p}|\mathbf{P}) := \bigoplus\limits_{i=1}^L \Phi_i(\mathbf{p} | \mathcal{E}(\mathbf{P})),
\end{equation}
where $\oplus$ denotes concatenation of activations from each layer of $\Phi$, and $L$ is the number of hidden layers. This feature serves as an implicit occupancy representation and provides rich geometric information about the geometric relationship between the queried point and the object surface.

In our implementation, $\Phi(\mathbf{p}|\mathcal{E}(\mathbf{P}))$ is a double-headed model that predicts both occupancy and signed distance, similar to \cite{huang2024implicit}. While occupancy 
prediction mainly captures features for points near contact regions, adding additional signed distance prediction helps capture features at points which are not in contact, which facilitates RL training when exploring non-contact regions. In our method, the NDF model is pretrained and kept fixed during RL training.
\subsection{Contact-Aware Dynamic Recovery}
\subsubsection{Contact-Aware Grasp Feature}
Recovering with an expected grasp primarily requires accurate modeling of finger–object contact. Beyond this, it is highly desirable for the contact representation to generalize across object geometries, as such generalization enables the robot policy to exhibit zero-shot generalization to unseen objects.


In this work, we leverage NDFs to characterize contact features for dexterous manipulation. By querying points on the hand, we can interpret the corresponding NDF features as indicators of contact: contact points are expected to lie near the decision boundary of the occupancy function, where the NDF occupancy prediction output transitions between \textit{inside} and \textit{outside} predictions.

While NDF provides per-point contact features, it does not directly characterize the contact features of a grasp. To address this, we predefine $K$ key points on the hand $\{ \mathbf{p}_{i}^{k_i} \}_{i=1}^K$, where $\mathbf{p}_{i}^{k_i}$ denotes the position of the $i$-th key point in the $k_i$-th link frame. We construct the raw grasp feature $G(\mathbf{q}, \mathbf{x}|\mathbf{P}) \in \mathbb{R}^{K \times n}$ by aggregating the NDF features of all $K$ key points: 
\begin{equation}
\begin{aligned}
    && G(\mathbf{q}, \mathbf{x}|\mathbf{P}) := [\boldsymbol{\phi}_1, \boldsymbol{\phi}_2, \cdots, \boldsymbol{\phi}_K ]^T \\
    && \boldsymbol{\phi}_i := f_{NDF}(T(\mathbf{x})f_{FK}(\mathbf{p}_i^{k_i}, \mathbf{q}, k_i) | \mathbf{P}),
    \label{equ:grasp_NDF}
\end{aligned}
\end{equation}
where $\mathbf{q}$ is the robot joint angles, $\mathbf{x} \in \mathrm{SE}(3)$ is the object pose. $f_{FK}$ denotes the forward kinematics function returning key point locations in the world frame, and $T(\mathbf{x})$ is the transformation from the world frame to the object frame. 

Because NDF features are extracted by concatenating multiple latent layers of a neural network, the raw grasp feature $G$ is high-dimensional and potentially redundant. 
We apply a shared projection matrix to compress these features into a lower-dimensional space, exploiting the structural similarity of NDF features across key points. This compressed representation is subsequently processed by a self-attention layer to further extract task-relevant contact relationships. The contact-aware grasp feature is defined as: 
\begin{equation}
    g(\mathbf{q}, \mathbf{x}|\mathbf{P}) := \text{Flatten}( G\mathbf{W}^T + \psi(\text{Attn}(G\mathbf{W}^T))),
    \label{equ:attention}
\end{equation}
where $\mathbf{W}$ is a learned projection matrix that compresses the raw NDF features into a lower-dimensional space, $\psi$ is an activation function, and $\text{Attn}$ is the self-attention operator. 

For key point selection, we sample points from each hand link rather than restricting them to the fingertips. We use the root of each link as the key point in our experiments. Although fingertips are often primarily involved in contact, non-contact parts also play a critical role in characterizing the grasp feature. For example, one of the major distinctions between a power grasp and a precision grasp is whether the palm is in contact with the object. 
In practice, we assign one key point for each link for computational efficiency. 
Moreover, it is not necessary to place key points strictly within potential contact regions, since NDF features encode not only whether a key point is in contact but also its distance from contact.

In summary, CADRE receives observations $\mathbf{o}_t$ and computes the contact-aware grasp feature $\mathbf{g} = g(\mathbf{q}, \mathbf{x}|\mathbf{P})$ at every time step. The input into the RL policy is the combination of observations and grasp features.
\subsubsection{Implicit Recovery Affordance} While contact-aware grasp features provide essential contact awareness, they do not inherently encourage recovery to task-appropriate states. To address this, we formulate an Implicit Recovery Affordance (IRA) that serves as a dense RL reward signal to assess state quality. To train the affordance function, we assume access to a dataset of primary task manipulation (e.g., screwdriver turning) performed on a \textbf{single} reference object: $\mathcal{D}=\{(\mathbf{q}_i, \mathbf{x}_i )\}_1^N$. The dataset consists of trajectories of robot and object states during successful task execution. Our objective is to learn an affordance function that generalizes zero-shot to novel object geometries, circumventing the tedious efforts of collecting object-specific demonstrations. We consider states that are in distribution (ID) of the demonstration dataset as desired recovery targets, while out-of-distribution (OOD) states are considered undesired. A primary challenge of training the affordance function is that NDF features are high-dimensional but reside on a much lower-dimensional manifold, as they are entirely parameterized by the low-dimensional kinematic state of the hand and object. Naively sampling negative examples in the NDF feature space yields feature vectors that do not correspond to any valid grasp configuration. Consequently, an unconstrained affordance model would trivially learn to distinguish between valid and invalid NDF features, rather than distinguishing between task-appropriate (ID) and task-inappropriate (OOD) grasps.
To overcome this issue, we apply Noise Contrastive Estimation(NCE)~\cite{gutmann2010noise}, generating negative data strictly within the valid grasp feature space. We first generate negative states $\{\mathbf{q}_i^-, \mathbf{x}_i^-\}$ by injecting Gaussian noise $\epsilon \in \mathcal{N}(0, \boldsymbol{\sigma})$ to the demonstration dataset. Because simple noise injection often results in unrealistic states with penetration, we pass the states through a simulator to resolve penetration.
\begin{equation}
    [\mathbf{q}_i^{-}, \mathbf{x}_i^{-}] = f_{sim}([\mathbf{q}_i, \mathbf{x}_i] + \epsilon), \quad \text{for } i = 1,2, \cdots, N
\end{equation}
Using these positive and negative datasets, we train a discriminator $f_{aff}$ to predict the likelihood that a queried state belongs to the target demonstration distribution. The network prediction serves as a score assessing the quality of the state for the primary manipulation task. To enable geometric generalization through contacts and prevent overfitting to the specific object poses and robot configurations from the demonstration with a single reference object, the discriminator takes only NDF grasp features as input: $r_{rec} = f_{aff}(G(\mathbf{q}, \mathbf{x}|\mathbf{P}))$. For network architecture, we use the same attention-based structure defined in Equation~\ref{equ:attention}, but utilize a separate set of weights for NDF processing. Furthermore, to provide an informative RL reward, the network must predict a smooth transition from OOD to ID states. We achieve this by applying Spectral Normalization~\cite{miyato2018spectral}, which bounds the network's Lipschitz constant.
\looseness=-1
\subsubsection{Reinforcement Learning for Dynamic Recovery}
We choose to use Proximal Policy Optimization (PPO)~\cite{schulman2017proximal} to optimize the dynamic recovery policy.  

\textbf{Reward Function}:
The reward function is defined as:
\begin{equation}
\begin{aligned}
    &r := r_{drop} + r_{obj\_v} + r_{contact} + r_{safety} + r_{rec} + r_{reg},\\
    &r_{reg} := r_{\dot{a}} + r_{torque} + r_{energy} + r_{obj\_pose} + r_{q}.
\end{aligned}
\end{equation}
We omit the weighting parameters and the time step subscript for simplicity.
$r_{drop} := -\ind_{drop}(\mathbf{x})$, where $\ind_{drop}$ is an indicator function that returns $1$ if the object's position exceeds a predefined workspace boundary. $r_{obj\_v} := \exp(-||\mathbf{v}||^2) + \exp(-||\boldsymbol{\omega}||^2)$ encourages low linear velocity $\mathbf{v}$ and angular velocity $\boldsymbol{\omega}$ of the object, which represents the objective of catching the object. We shape $r_{obj\_v}$ with an exponential function as the velocities can yield excessively large magnitudes during training (e.g., the object is falling fast). The exponential function helps bound the reward and leads to more stable training.  $r_{reg}$ is the regularization term, in which $ r_{\dot{\mathbf{a}}}:= -||\mathbf{a}_{t} - \mathbf{a}_{t-1}||^2$, $r_{toruqe} := -||\tau||^2$, and $r_{energy} := -||\tau^T\dot{q}||^2$ regularizes non-smooth actions, large torque and large energy consumption, respectively. $\mathbf{\tau}$ is computed from the joint position controller output. 
$r_{q} := \exp(-||\mathbf{q} - \mathbf{\hat{q}}||^2)$ and $r_{obj\_pose}:= \exp(-f_{pos}(\mathbf{x}, \mathbf{\hat{x}})) + \exp(-f_{orn}(\mathbf{x}, \mathbf{\hat{x}}))$ rewards the robot for staying near the nominal robot configuration $\hat{\mathbf{q}}$ and object pose $\hat{\mathbf{x}}$, where $\hat{\mathbf{q}}$ and $\hat{\mathbf{x}}$ are defined as the mean of the environment's initial state distribution during RL training. 
$f_{pos}$ and $f_{orn}$ computes the position and orientation difference, respectively. $r_{contact} := \ind_{contact}(\mathbf{q}, \mathbf{x})$ rewards the robot for achieving task-relevant contact with the object. The desired task-relevant contact can be obtained from the primary task demonstration dataset (see Sec.~\ref{sec:exp_setup}). $r_{safety}:=-\ind_{table}(\mathbf{q})$ penalizes the unsafe behavior of robot contacting the table.
\vspace{-0.2cm}   
\section{Experiments}
\vspace{-0.2cm}
\label{sec:exp}
We evaluate our method on two tasks: screwdriver recovery and hole-on-peg (inverted peg-in-hole) recovery. We aim to design our experiments to answer (1) whether incorporating implicit contact features can facilitate RL training and improve catching performance; (2) whether defining IRA facilitates the recovery performance, (3) whether CADRE generalizes to unseen objects sampled from a different distribution; and (4) whether CADRE can be deployed on robot hardware. 
\vspace{-0.2cm}
\subsection{Simulation Experiment Setup}
\vspace{-0.2cm}
\label{sec:exp_setup}
We use IsaacSim to simulate the recovery task. Our robot consists of an Allegro Hand mounted on a 7-DoF KUKA iiwa arm. We use PPO~\cite{schulman2017proximal}, implemented from RL games~\cite{rl-games2021}, to optimize the recovery policy. To obtain the full point cloud of the object, we assume access to its geometry (as noted in Sec.~\ref{sec:problem_statement}), since capturing a complete point cloud with a camera is impractical. We uniformly sample points from the object surface and transform them according to the object pose to generate the full point cloud observation. The same approach is used in our hardware experiments.

\textbf{Object generation:} During RL training, we sample 50 objects from a predefined shape distribution. The objects are generated using the same geometric parameterization method, but with variations in their size parameters. 
During training for the screwdriver task we add five real screwdrivers to support our real-world experiments.
For evaluation, we consider both ID and OOD objects. ID objects consist of 5 new unseen objects sampled from the same distribution used for training, while OOD objects are sampled from a different distribution as a more challenging test for generalization (see appendix).\looseness=-1
The NDF model $f_{NDF}$ is trained with objects sampled from the same distribution for RL training. Thus, OOD objects for the RL policy will also be OOD for the NDF model. 
\looseness=-1

\textbf{Evaluation metrics:} 
We consider the following metrics:
(1) Catch success rate: a catch is considered successful if the object remains within a predefined bounding box throughout the episode; (2) Number of desired contacts; (3) Number of undesired contacts; (4) the primary task performance. While Metric (1) is the baseline requirement for successful recovery, our method emphasizes achieving task-appropriate contacts, which is evaluated through Metrics (2)–(4). Metrics (2) and (3) are task-specific and detailed in Sec.~\ref{sec:desired contact}. Metrics (2)-(4) are \textbf{only calculated for successful trials}. 
Metric (4) evaluates downstream task performance by resuming a screwdriver turning or hole-on-peg task directly from the final recovered states (Sec.~\ref{sec:setup_screwdriver}, \ref{sec:setup_peg_on_hole}).
\looseness=-1
\subsubsection{Desired Contact Extraction}
\label{sec:desired contact}
To extract the desired task-relevant contact, we compute the finger–object distance in the demonstration dataset. Fingers with a distance below $2~\text{cm}$ for more than 80 \% of the time steps are classified as task-relevant contacts, corresponding to the index, middle, and thumb fingers for the screwdriver task and all four fingers for the hole-on-peg task. The undesired contact is defined as the contact between all other links of the hand with the screwdriver. 
\looseness=-1
\subsubsection{Screwdriver Recovery}
\label{sec:setup_screwdriver}
We set up the recovery task in a screwdriver turning scenario with a precision grasp.  
Screwdriver turning with a precision grasp has been widely studied~\cite{naughton2024respilot, tang2024robotic, yang2024multi, kumar2024diffusion}. A precision grasp is preferred, as opposed to a power grasp, as the choice of robot-screwdriver contact significantly affects the turning performance, highlighting the importance of contact reasoning. Specifically, we follow the setup presented in \cite{yang2024multi} (See Fig.~\ref{fig:screwdriver_setup}), where the index finger contacts the top of the screwdriver, while the thumb and middle finger form an antipodal grasp on the handle. \looseness=-1


We model the screwdriver as two connected cylinders: one for the handle and one for the shaft. 
OOD screwdrivers have approximately half the length of the ID screwdrivers.

To evaluate whether the recovered states are favorable for screwdriver turning, we follow the setup from \cite{yang2024multi}, attempting to turn the screwdriver $60^\circ$. We assume there exists a motion planning algorithm to mate the screwdriver with the screw and reorient it perfectly upright. In practice, we record the final state of recovery, keep the joint angles of the Allegro hand and the relative transformation between the hand and the screwdriver fixed, but we transform both the hand and the screwdriver so that the screwdriver is upright. We evaluate the turning performance via the turning drop rate and the object orientation difference between the final turning state and the desired state.
We consider the screwdriver dropped if its Euler angles exceed a predefined threshold. Additionally, the same screwdriver turning algorithm is also used to generate demonstrations for training the IRA function.
\subsubsection{Hole-on-Peg Recovery}
\label{sec:setup_peg_on_hole}
\begin{figure}
\vspace{0.15cm}
  \centering
  \begin{subfigure}[t]{0.16\textwidth}
    \centering
    \includegraphics[width=\linewidth]{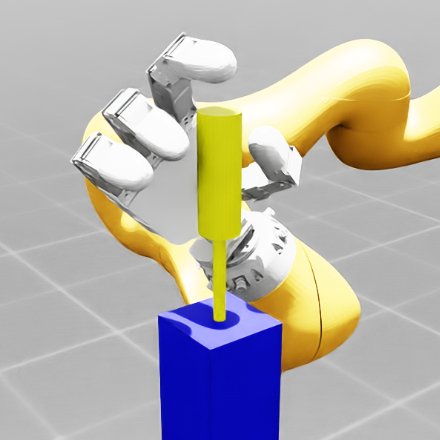}
    \caption{Screwdriver setup}
    \label{fig:screwdriver_setup}
  \end{subfigure}
  \begin{subfigure}[t]{0.16\textwidth}
    \centering
    \includegraphics[width=\linewidth]{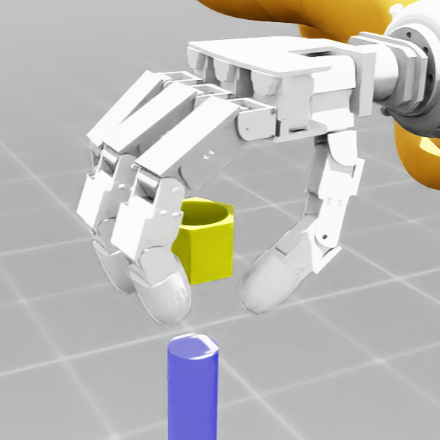}
    \caption{Hole-on-peg setup}
    \label{fig:peg_on_hole_setup}
  \end{subfigure}
  \vspace{-0.7cm}
\end{figure}

We consider recovery for the hole-on-peg (inverted peg-in-hole) task (See Fig.~\ref{fig:peg_on_hole_setup}). During mating, the robot can easily drop the socket as it might involve forceful contacts and high uncertainty, especially if tactile sensors are not available. In hole-on-peg recovery tasks, successful recovery requires the robot to catch the socket with a specific grasp: all fingertips should make contact with the side of the socket while avoiding its top and bottom. Contacts on the top or bottom will block the socket's central hole, making it impossible to mate with the peg.
\looseness=-1

Additionally, there are cases where the robot appears to stabilize the socket by using the external support from the peg. For instance, the robot might push the socket towards the peg while the grasp itself is not stable without the support from the socket. We penalize such behaviors in the contact reward: $r_{contact} := r_{robot\_socket} - r_{socket\_peg}$, where $r_{robot\_socket}$ returns the number of desired contacts and $r_{socket\_peg}$ is the indicator function for socket-peg contact. 

The socket is modeled as a hexagonal prism with a cylindrical hole, and the peg is modeled as a cylinder. 
OOD sockets have about half the thickness of the ID sockets, requiring more precise finger control to catch. 

A recovery is considered successful if the socket is not dropped, and the socket does not contact the peg, as the robot must stably grasp the socket without the peg's support.

To evaluate grasp quality for the primary hole-on-peg task, we design a simplified experiment. Specifically, we use a peg with a bit smaller radius to allow more clearance, and consider the task successful if the peg passes through the socket's center hole. 
This simplification primarily tests whether the robot’s fingers obstruct the socket’s center hole, a necessary condition for successful hole-on-peg execution. During the experiment, we assume the finger configurations remain fixed and use motion planning to move the arm. The same motion planning algorithm is also used to generate demonstrations for training the IRA function.

\subsubsection{Baselines and Ablation}
Baselines and ablations evaluate the impact of (1) alternative contact representations and (2) the IRA on recovery performance. All methods use the same reward functions and RL training setup except the IRA function: (1) \textbf{Ablation: object pose observations (CADRE-No NDF):} it directly takes the observation $\mathbf{o}_t$ defined in Sec.~\ref{sec:problem_statement} as policy input. This method does not have access to object geometry; the only object-related information available is pose and velocity. 
(2) \textbf{Ablation: CADRE without IRA (CADRE-No IRA):} this method removes the IRA function from the RL reward function to isolate its impact on recovery performance. (3) \textbf{Baseline: point cloud and obj pose observations (PC + Pose):} Reinforcement learning with point cloud input has been widely used in dexterous manipulation~\cite{huang2021generalization, wu2023learning, bao2023dexart}. Similar to these methods, we add the object's point cloud $\mathbf{P}_t$ into the aforementioned observation space: $\mathbf{o}'_t:= \{\mathbf{q}_t, \mathbf{x}_t, \mathbf{v}_t, \mathbf{P}_t \}$.We choose not to use the point cloud obtained from a depth camera to maintain the same input as our method. This method has access to the object's geometry but does not explicitly reason about contact features. PointNet~\cite{qi2017pointnet} is used to extract features from the point cloud. (4) \textbf{Baseline: DexPoint~\cite{qi2017pointnet}:} Similar to our method, DexPoint leverages contact information to improve generalization across object geometries. Its key component is augmenting the observation space $\mathbf{o}_t$ with an \textbf{additional} imagined point cloud of the robot. To implement DexPoint, we render a point cloud for each fingertip and the palm, and concatenate it with the object point cloud. Unlike the original DexPoint paper, we do not use the observed point cloud, i.e., the depth-camera point cloud, since other methods in our experiments do not have access to such observations. PointNet is also used to encode the point cloud. Note that since we also add IRA into DexPoint training, and our implementation includes more components than the original architecture. (5) \textbf{Baseline: CADRE with an energy-based model(EBM):} recent advances have demonstrated the efficacy of energy-based models (EBMs) for data distribution modeling and OOD detection. To demonstrate that our NCE formulation is better suited for high-dimensional NDF features, we replace it with EBM from Du and Mordatch~\cite{du2019implicit} to model the IRA. 

\textbf{Affordance Inputs:} We adapt the IRA network input to match each baseline's representation, supplementing with $\mathbf{q}_t$ and $\mathbf{x}_t$ when implicit contact features are absent: CADRE-No NDF uses $\{\mathbf{q}_t, \mathbf{x}_t \}$, PC + Pose uses $\{\mathbf{q}_t, \mathbf{x}_t, \mathbf{P}_t \}$, and DexPoint uses robot and object point clouds.
\vspace{-0.2cm}
\subsection{Experiment Results}
\vspace{-0.2cm}
\begin{table*}
\small
\setlength{\tabcolsep}{2pt} 
    \centering
    \vspace{0.2cm}
    \begin{tabular}{|c|c|c|c|c|c|c V{3} c|c|c|c|c|c|}
    \hline
     & \multicolumn{3}{c|}{Screwdriver ID} & \multicolumn{3}{c V{3}}{hole-on-peg ID} & \multicolumn{3}{c|}{Screwdriver OOD} & \multicolumn{3}{c|}{hole-on-peg OOD}\\
     \hline
         method & success$\uparrow$ & $C_{des}\uparrow$ & $C_{und}\downarrow$ & success$\uparrow$ & $C_{des}\uparrow$ & $C_{und}\downarrow$  & success$\uparrow$ & $C_{des}\uparrow$ & $C_{und}\downarrow$  & success$\uparrow$ & $C_{des}\uparrow$ & $C_{und}\downarrow$ \\
        \hline
     CADRE  & \textbf{100.00}\% & \textbf{2.96} & \textbf{0.03} & 99.33\% & \textbf{3.90} & 0.05 &\textbf{99.87}\% & 1.55 & \textbf{0.51} & 91.20\% & \textbf{3.03} & 0.07\\
     \hline
     PC + Pose & \textbf{100.00}\% &1.95 & 1.93 & 71.47\%& 1.30 & 0.05 &80.93\% & 1.15 & 0.70 & 60.53 \% & 0.87 & 0.28\\
     \hline
     DexPoint &99.73\% & 2.39 & 0.65 & 91.73\% & 3.59 & 0.31 &95.07\% & 1.70 & 0.54 & 64.27\% & 2.28 & 0.04\\
     \hline
      CADRE-EBM & \textbf{100.00}\% & 2.32 & 1.47 & \textbf{99.47}\% & 2.64& 0.95 &98.13\% & \textbf{1.73} & 1.33 & \textbf{93.33}\% & 1.64 & 0.93\\
     \hline
      CADRE-No NDF & \textbf{100.00}\% & 2.22 & 1.17 & 56.53\% & 1.64 & \textbf{0.02} &72.93\% & 1.15 & 0.56 & 39.87\% & 1.37 & \textbf{0.00}\\
     \hline
     CADRE-No IRA & \textbf{100.00}\% & 2.63 & 1.38 & 99.07\% & 3.23 & 0.54 &92.93\% & 1.45 & 1.26 & 90.04\% &2.36 & 0.51 \\
     \hline
    \end{tabular}
    \caption{Unseen screwdriver recovery and Hole-on-Peg recovery. $C_{des}$ and $C_{und}$ denote the number of desired and undesired contacts, respectively. Those contact metrics are highly related to the downstream task performance. Results are calculated for successful trials only.}
    \vspace{-0.3cm}
    \label{tab:screwdriver_and_socket}
\end{table*}

Each method is evaluated for 50 trials per object. The average performance across all seeds is reported in Table~\ref{tab:screwdriver_and_socket}.

\textbf{Screwdriver recovery:} all methods achieve almost 100\% success rates on ID objects. This is reasonable as dropping the object is penalized during RL training. However, CADRE performs better on contact metrics. This suggests that baselines often use power grasps rather than the desired precision grasps to catch the screwdriver. 
On OOD objects, CADRE, the CADRE-EBM, CADRE-No IRA and DexPoint still maintain similar performance, while the other methods exhibit a more significant performance drop. However, we will demonstrate in Table~\ref{tab:primary_task_eval} that grasp poses of DexPoint and CADRE variants are not well-suited for the primary screwdriver-turning task.

For the downstream evaluation (see Sec.~\ref{sec:setup_screwdriver}), we randomly sample 5 recovery trials per object per seed because the turning algorithm~\cite{yang2024multi} is computationally expensive.
The turning results are shown in Table~\ref{tab:primary_task_eval}. According to the results, CADRE has demonstrated a much smaller distance to goal in ID scenarios and a lower drop rate in OOD scenarios. Its distance to goal is slightly higher. This is a direct consequence of its low drop rate: in difficult scenarios where baseline methods drop the screwdriver, CADRE manages a partial turn. This results in a larger average distance-to-goal that reflects the inherent difficulty of these edge-case setups.


\begin{table*}
    \centering
    \small
    \setlength{\tabcolsep}{2pt} 
    \begin{tabular}{|c|c|c|c|c V{3} c|c|c|c|c|}
    \hline
         & \multicolumn{2}{c|}{Screw Driver Turning ID} & \multicolumn{2}{c V{3}}{Screw Driver Turning OOD} & \multicolumn{2}{c|}{hole-on-peg ID} & \multicolumn{2}{c|}{hole-on-peg OOD} \\
         \hline
         & drop rate$\downarrow$& \makecell{Dist. to Goal$\downarrow$ \\(no-drop trials)} & drop rate$\downarrow$& \makecell{Dist. to Goal$\downarrow$ \\(no-drop trials)} & valid succ$\uparrow$ & overall succ$\uparrow$& valid succ$\uparrow$ & overall succ$\uparrow$\\
         \hline
        CADRE & \textbf{0}\% & $\textbf{16.79}^\circ \pm 19.07^\circ$ & \textbf{28}\% & $40.39^\circ \pm 20.07^\circ$ & \textbf{94.36}\% & \textbf{93.73}\% & 76.61\% & \textbf{69.87}\%\\
        \hline
        PC + Pose & 2.67\% & $49.30^\circ \pm 15.23^\circ$ & 57.33\% & $42.43^\circ \pm 19.69^\circ$ & 38.06\% & 27.02\% & 38.77\% & 23.47\%\\
        \hline
        DexPoint & 1.33\% & $42.17^\circ \pm 19.93^\circ$ & 52\% & $32.47^\circ \pm 25.74^\circ$ & 89.83\% & 82.40\% & 81.12\% & 52.13\%\\
        \hline
        CADRE-EBM & 8.00\% & $47.62^\circ \pm 17.23^\circ$ & 45.33\% & $\textbf{28.14}^\circ \pm 17.23^\circ$ & 57.51\% & 57.20\% & 37.31\% & 35.20\%\\
        \hline
        CADRE-No NDF & 1.33\% & $36.75 ^\circ \pm 16.34^\circ$ & 53.33\% & $ 36.11^\circ \pm 18.66^\circ$ & 77.59\% & 43.87\% & \textbf{86.62}\% & 34.53\%\\
        \hline
        CADRE-No IRA & 2.67\% & $33.78^\circ \pm 21.27^\circ$ & 34.67\% & $36.53^\circ \pm 18.94^\circ$ & 70.66\% & 70.00\% & 57.23\% & 51.73\% \\
        \hline
    \end{tabular}
    \caption{Downstream task performance initialized from recovered states. Dist. to Goal is the screwdriver orientation difference between the final state of turning and the desired target of turning $60^\circ$. hole-on-peg includes two success rates: valid success rate$(\# \text{successful mating} / \# \text{successful cathing})$, and the overall success rate$(\# \text{successful mating} / \# \text{trials})$, where the number of trials also includes failed catching attempts.}
    \label{tab:primary_task_eval}
    \vspace{-0.7cm}
\end{table*}

\textbf{hole-on-peg recovery:} 
We have also observed similar results in the hole-on-peg task, where CADRE outperforms the baselines in both ID and OOD scenarios. 
In the downstream hole-on-peg task, we sample 50 recovery trials per object for evaluation. Although some baselines achieve higher mating success rates, CADRE attains a high catching success rate, resulting in a higher overall success rate from catching to mating. In particular, baselines tend to successfully catch the socket mainly in simpler catching scenarios, which makes it easier for them to achieve higher mating performance.

Furthermore, CADRE-EBM's underperformance compared to CADRE-No IRA suggests that the EBM from Du and Mordatch~\cite{du2019implicit} struggles to learn effective affordances in this high-dimensional feature space.
\vspace{-0.2cm}
\subsection{Hardware Experiments}
\vspace{-0.1cm}
The hardware experiments evaluate whether the highly dynamic recovery behaviors successfully transfer from simulation to reality. We only focus on the screwdriver recovery task. 
We use an unseen real screwdriver that falls within the training distribution and the Vicon motion capture system to estimate the state of the screwdriver for closed-loop control.

\subsubsection{Finetune policies to reduce contact forces}
The RL policies trained in the simulation experiments are not designed to explicitly avoid high contact forces. Directly deploying these policies in the real world is unsafe and highly vulnerable to sim-to-real gaps, as larger forces typically induce faster movements that magnify modeling errors. Therefore, we update the contact reward function
$r'_{contact} := \sum_i\mathbf{1}_{contact}^i \exp(-\max(||\mathbf{f}_i|| - f_{min},0))$,
where $\mathbf{f}_i$ is the contact force between the $i$th fingertip and the object, and $\mathbf{1}_{contact}^i$ is the indicator function of whether the $i$th fingertip is in contact. $ f_{min}$ is the predefined minimum threshold. The real-world policy is finetuned from the simulation policy that achieved the highest catching success rate. While retraining from scratch is an alternative, optimizing for low contact forces is challenging in our dynamic recovery scenarios. The pretraining-finetuning pipeline effectively acts as a curriculum, yielding a higher training efficiency. Each policy is finetuned for 7000 additional epochs in sim. 
\subsubsection{Experiment Setup} 
Consistently resetting to a pre-fall initial state is challenging. As the error detection is not the focus of this work, we manually create such a falling scenario: we first manually position the screwdriver so that the robot can grasp it. To initiate the drop, we command the robot to open its hand for a short, fixed duration. In our experiment, this drop duration is set to be 0.17 seconds. Note that the screwdriver falls to a horizontal resting position in approximately 0.4 seconds, after which it becomes irrecoverable. The recovery policy is then triggered. To eliminate bias from manually setting the screwdriver's initial pose, we anonymize the methods during each trial: the method under evaluation is randomly selected and not revealed to the experimenter until the end of this trial.
In our experiments, we use a real screwdriver but add friction tape, as its material—optimized for human grip—does not provide sufficient friction for the robot.\looseness=-1
\subsubsection{Experiment Results}
Results are shown in Table~\ref{tab:hardware}.
Our method demonstrates a better performance in the hardware experiments. CADRE-No IRA performs significantly worse in hardware than in sim. This is likely because it lacks an explicit understanding of grasp quality. The finetuning procedure encourages a policy that exerts minimal force; however, without the affordance reward ensuring appropriate finger placement, the policy simply learns to apply \textbf{overly} low forces. Consequently, the grasp becomes highly vulnerable to the sim-to-real gap, where minor modeling errors easily break contact. The control frequency is set to 30 Hz. Despite intensive NDF computations, a single CADRE inference step takes only 4.84 ms. Nevertheless, the results still reveal a noticeable sim-to-real gap, indicating that further improvement is needed before deployment in production. Potential improvements include reducing action jerk and obtaining more accurate estimates of joint friction and robot inertia. We leave closing the sim-to-real gap for highly dynamic systems as future work. \looseness=-1
\begin{table}
    \centering
    \setlength{\tabcolsep}{2pt} 
    \begin{tabular}{|c|c|c|c|}
    \hline
         &  succ rate $\uparrow$ & $C_{des}\text{(succ only)}\uparrow$ & $C_{und}\text{(succ only)}\downarrow$\\
         \hline
       CADRE  & \textbf{86.7\%} & \textbf{2.3} & \textbf{1.0} \\
       \hline
       CADRE-No IRA & 0.0\% & \diagbox{}{} & \diagbox{}{}\\
       \hline
       DexPoint & 73.3\% & 1.9 & 1.5\\
       \hline
    \end{tabular}
    \caption{Each method is evaluated for 15 hardware trials. 
    }
    \vspace{-0.4cm}
    \label{tab:hardware}
\end{table}
\vspace{-0.2cm}
\section{Discussion}
\vspace{-0.2cm}
While CADRE demonstrates the ability to generalize across geometries, the underlying NDF structure only focuses on geometry for generalization. However, geometric variations also lead to different dynamics. For example, a grasp on an unseen object sharing similar grasp features with a force closure grasp on a known object might not result in force closure. 
\update{Additionally, while CADRE successfully catches a falling object in our hardware experiment, deploying such systems in industry remains challenging due to the large sim-to-real gap of dynamic manipulation. Specifically, errors in robot mass, inertia, and joint friction estimation, and actuator dynamics, which usually do not affect quasi-static manipulation, can severely degrade performance on highly-dynamic recovery tasks. Modeling dynamics discrepancies or incorporating tactile feedback could achieve desired contacts more precisely, enabling recovery from more challenging failure scenarios. CADRE also currently assumes access to full object geometry. Future work will focus on distilling CADRE into reactive policies that utilize onboard vision and tactile feedback. }\looseness=-1 

Despite the above limitations, our results suggest the broader potential of using implicit contact representations in an RL framework for contact-rich manipulation. 
\vspace{-0.1cm}
\section{Conclusion}
\vspace{-0.1CM}
\label{sec:conclusion}
In this work, we focus on the problem of recovering from catastrophic failure in dexterous manipulation--specifically, recovering from dropping objects and returning to states favorable for resuming the primary manipulation task. We propose CADRE, an RL framework that utilizes an implicit contact representation derived from NDFs. 

We evaluate CADRE on screwdriver and hole-on-peg recovery tasks. Our experiments demonstrate that simply providing point clouds as object geometry observations for RL is insufficient for learning effective dynamic recovery in contact-rich scenarios. In contrast, CADRE leverages implicit contact features to improve training efficiency, recovery quality, and generalization to unseen object geometries. 
\vspace{-0.1cm}
\bibliographystyle{IEEEtran}
\bibliography{reference}  

@article{yang2025contactsdf,
  title={Contactsdf: Signed distance functions as multi-contact models for dexterous manipulation},
  author={Yang, Wen and Jin, Wanxin},
  journal={RA-L},
  year={2025},
  publisher={IEEE}
}

@article{wang2024neural,
  title={Neural attention field: Emerging point relevance in 3d scenes for one-shot dexterous grasping},
  author={Wang, Qianxu and Deng, Congyue and Lum, Tyler Ga Wei and Chen, Yuanpei and Yang, Yaodong and Bohg, Jeannette and Zhu, Yixin and Guibas, Leonidas},
  journal={arXiv preprint arXiv:2410.23039},
  year={2024}
}

@article{wei2024mathcal,
  title={$D (R, O)$ Grasp: A Unified Representation of Robot and Object Interaction for Cross-Embodiment Dexterous Grasping},
  author={Wei, Zhenyu and Xu, Zhixuan and Guo, Jingxiang and Hou, Yiwen and Gao, Chongkai and Cai, Zhehao and Luo, Jiayu and Shao, Lin},
  journal={arXiv preprint arXiv:2410.01702},
  year={2024}
}

@article{du2019implicit,
  title={Implicit generation and modeling with energy based models},
  author={Du, Yilun and Mordatch, Igor},
  journal={NeurIPS},
  volume={32},
  year={2019}
}

@article{miyato2018spectral,
  title={Spectral normalization for generative adversarial networks},
  author={Miyato, Takeru and Kataoka, Toshiki and Koyama, Masanori and Yoshida, Yuichi},
  journal={arXiv preprint arXiv:1802.05957},
  year={2018}
}

@inproceedings{gutmann2010noise,
  title={Noise-contrastive estimation: A new estimation principle for unnormalized statistical models},
  author={Gutmann, Michael and Hyv{\"a}rinen, Aapo},
  booktitle={Proceedings of the thirteenth international conference on artificial intelligence and statistics},
  pages={297--304},
  year={2010},
  organization={JMLR Workshop and Conference Proceedings}
}

@article{feix2015grasp,
  title={The grasp taxonomy of human grasp types},
  author={Feix, Thomas and Romero, Javier and Schmiedmayer, Heinz-Bodo and Dollar, Aaron M and Kragic, Danica},
  journal={IEEE Transactions on human-machine systems},
  volume={46},
  number={1},
  pages={66--77},
  year={2015},
  publisher={IEEE}
}

@article{kim2014catching,
  title={Catching objects in flight},
  author={Kim, Seungsu and Shukla, Ashwini and Billard, Aude},
  journal={IEEE Transactions on Robotics},
  volume={30},
  number={5},
  pages={1049--1065},
  year={2014},
  publisher={IEEE}
}

@inproceedings{simeonov2022neural,
  title={Neural descriptor fields: Se (3)-equivariant object representations for manipulation},
  author={Simeonov, Anthony and Du, Yilun and Tagliasacchi, Andrea and Tenenbaum, Joshua B and Rodriguez, Alberto and Agrawal, Pulkit and Sitzmann, Vincent},
  booktitle={2022 ICRA},
  pages={6394--6400},
  year={2022},
  organization={IEEE}
}

@inproceedings{simeonov2023se,
  title={Se (3)-equivariant relational rearrangement with neural descriptor fields},
  author={Simeonov, Anthony and Du, Yilun and Lin, Yen-Chen and Garcia, Alberto Rodriguez and Kaelbling, Leslie Pack and Lozano-P{\'e}rez, Tom{\'a}s and Agrawal, Pulkit},
  booktitle={CoRL},
  pages={835--846},
  year={2023},
  organization={PMLR}
}

@article{schulman2017proximal,
  title={Proximal policy optimization algorithms},
  author={Schulman, John and Wolski, Filip and Dhariwal, Prafulla and Radford, Alec and Klimov, Oleg},
  journal={arXiv preprint arXiv:1707.06347},
  year={2017}
}

@inproceedings{tobin2017domain,
  title={Domain randomization for transferring deep neural networks from simulation to the real world},
  author={Tobin, Josh and Fong, Rachel and Ray, Alex and Schneider, Jonas and Zaremba, Wojciech and Abbeel, Pieter},
  booktitle={2017 IROS},
  pages={23--30},
  year={2017},
  organization={IEEE}
}

@article{naughton2024respilot,
  title={ResPilot: Teleoperated Finger Gaiting via Gaussian Process Residual Learning},
  author={Naughton, Patrick and Cui, Jinda and Patel, Karankumar and Iba, Soshi},
  journal={arXiv preprint arXiv:2409.09140},
  year={2024}
}

@inproceedings{tang2024robotic,
  title={Robotic manipulation of hand tools: The case of screwdriving},
  author={Tang, Ling and Jia, Yan-Bin and Xue, Yuechuan},
  booktitle={2024 ICRA},
  pages={13883--13890},
  year={2024},
  organization={IEEE}
}

@article{yang2024multi,
  title={Multi-finger Manipulation via Trajectory Optimization with Differentiable Rolling and Geometric Constraints},
  author={Yang, Fan and Power, Thomas and Marinovic, Sergio Aguilera and Iba, Soshi and Zarrin, Rana Soltani and Berenson, Dmitry},
  journal={arXiv preprint arXiv:2408.13229},
  year={2024}
}

@article{kumar2024diffusion,
  title={Diffusion-Informed Probabilistic Contact Search for Multi-Finger Manipulation},
  author={Kumar, Abhinav and Power, Thomas and Yang, Fan and Marinovic, Sergio Aguilera and Iba, Soshi and Zarrin, Rana Soltani and Berenson, Dmitry},
  journal={arXiv preprint arXiv:2410.00841},
  year={2024}
}

@misc{rl-games2021,
title = {rl-games: A High-performance Framework for Reinforcement Learning},
author = {Makoviichuk, Denys and Makoviychuk, Viktor},
month = {May},
year = {2021},
publisher = {GitHub},
journal = {GitHub repository},
howpublished = {\url{https://github.com/Denys88/rl_games}},
}

@inproceedings{qin2023dexpoint,
  title={Dexpoint: Generalizable point cloud reinforcement learning for sim-to-real dexterous manipulation},
  author={Qin, Yuzhe and Huang, Binghao and Yin, Zhao-Heng and Su, Hao and Wang, Xiaolong},
  booktitle={CoRL},
  pages={594--605},
  year={2023},
  organization={PMLR}
}

@article{huang2021generalization,
  title={Generalization in dexterous manipulation via geometry-aware multi-task learning},
  author={Huang, Wenlong and Mordatch, Igor and Abbeel, Pieter and Pathak, Deepak},
  journal={arXiv preprint arXiv:2111.03062},
  year={2021}
}

@inproceedings{wu2023learning,
  title={Learning generalizable dexterous manipulation from human grasp affordance},
  author={Wu, Yueh-Hua and Wang, Jiashun and Wang, Xiaolong},
  booktitle={Conference on Robot Learning},
  pages={618--629},
  year={2023},
  organization={PMLR}
}

@inproceedings{bao2023dexart,
  title={Dexart: Benchmarking generalizable dexterous manipulation with articulated objects},
  author={Bao, Chen and Xu, Helin and Qin, Yuzhe and Wang, Xiaolong},
  booktitle={CVPR},
  pages={21190--21200},
  year={2023}
}

@article{huang2024implicit,
  title={Implicit Contact Diffuser: Sequential Contact Reasoning with Latent Point Cloud Diffusion},
  author={Huang, Zixuan and He, Yinong and Lin, Yating and Berenson, Dmitry},
  journal={arXiv preprint arXiv:2410.16571},
  year={2024}
}

@article{zeng2020tossingbot,
  title={Tossingbot: Learning to throw arbitrary objects with residual physics},
  author={Zeng, Andy and Song, Shuran and Lee, Johnny and Rodriguez, Alberto and Funkhouser, Thomas},
  journal={IEEE Transactions on Robotics},
  volume={36},
  number={4},
  pages={1307--1319},
  year={2020},
  publisher={IEEE}
}

@inproceedings{ishihara2006dynamic,
  title={Dynamic pen spinning using a high-speed multifingered hand with high-speed tactile sensor},
  author={Ishihara, Tatsuya and Namiki, Akio and Ishikawa, Masatoshi and Shimojo, Makoto},
  booktitle={2006 6th IEEE-RAS International Conference on Humanoid Robots},
  pages={258--263},
  year={2006},
  organization={IEEE}
}

@inproceedings{namiki2003development,
  title={Development of a high-speed multifingered hand system and its application to catching},
  author={Namiki, Akio and Imai, Yoshiro and Ishikawa, Masatoshi and Kaneko, Makoto},
  booktitle={ROS 2003},
  volume={3},
  pages={2666--2671},
  year={2003},
  organization={IEEE}
}

@article{wang2024caging,
  title={Caging in Time: A Framework for Robust Object Manipulation under Uncertainties and Limited Robot Perception},
  author={Wang, Gaotian and Ren, Kejia and Morgan, Andrew S and Hang, Kaiyu},
  journal={arXiv preprint arXiv:2410.16481},
  year={2024}
}

@article{yang2024dynamic,
  title={Dynamic on-palm manipulation via controlled sliding},
  author={Yang, William and Posa, Michael},
  journal={arXiv preprint arXiv:2405.08731},
  year={2024}
}

@inproceedings{hou2020robust,
  title={Robust planar dynamic pivoting by regulating inertial and grip forces},
  author={Hou, Yifan and Jia, Zhenzhong and Johnson, Aaron M and Mason, Matthew T},
  booktitle={Algorithmic Foundations of Robotics XII: Proceedings of the Twelfth Workshop on the Algorithmic Foundations of Robotics},

  year={2020},
  organization={Springer}
}

@article{salehian2016dynamical,
  title={A dynamical system approach for softly catching a flying object: Theory and experiment},
  author={Salehian, Seyed Sina Mirrazavi and Khoramshahi, Mahdi and Billard, Aude},
  journal={IEEE T-RO},
  volume={32},
  number={2},
  pages={462--471},
  year={2016},
  publisher={IEEE}
}

@article{zhang2024catch,
  title={Catch it! learning to catch in flight with mobile dexterous hands},
  author={Zhang, Yuanhang and Liang, Tianhai and Chen, Zhenyang and Ze, Yanjie and Xu, Huazhe},
  journal={arXiv preprint arXiv:2409.10319},
  year={2024}
}

@article{huang2023dynamic,
  title={Dynamic handover: Throw and catch with bimanual hands},
  author={Huang, Binghao and Chen, Yuanpei and Wang, Tianyu and Qin, Yuzhe and Yang, Yaodong and Atanasov, Nikolay and Wang, Xiaolong},
  journal={arXiv preprint arXiv:2309.05655},
  year={2023}
}

@article{lan2023dexcatch,
  title={Dexcatch: Learning to catch arbitrary objects with dexterous hands},
  author={Lan, Fengbo and Wang, Shengjie and Zhang, Yunzhe and Xu, Haotian and Oseni, Oluwatosin and Zhang, Ziye and Gao, Yang and Zhang, Tao},
  journal={arXiv preprint arXiv:2310.08809},
  year={2023}
}

@inproceedings{charlesworth2021solving,
  title={Solving challenging dexterous manipulation tasks with trajectory optimisation and reinforcement learning},
  author={Charlesworth, Henry J and Montana, Giovanni},
  booktitle={ICML},
  pages={1496--1506},
  year={2021},
  organization={PMLR}
}

@article{driess2022reinforcement,
  title={Reinforcement learning with neural radiance fields},
  author={Driess, Danny and Schubert, Ingmar and Florence, Pete and Li, Yunzhu and Toussaint, Marc},
  journal={Advances in Neural Information Processing Systems},
  volume={35},
  pages={16931--16945},
  year={2022}
}

@article{mildenhall2021nerf,
  title={Nerf: Representing scenes as neural radiance fields for view synthesis},
  author={Mildenhall, Ben and Srinivasan, Pratul P and Tancik, Matthew and Barron, Jonathan T and Ramamoorthi, Ravi and Ng, Ren},
  journal={Communications of the ACM},
  volume={65},
  number={1},
  pages={99--106},
  year={2021},
  publisher={ACM New York, NY, USA}
}

@inproceedings{wu2023daydreamer,
  title={Daydreamer: World models for physical robot learning},
  author={Wu, Philipp and Escontrela, Alejandro and Hafner, Danijar and Abbeel, Pieter and Goldberg, Ken},
  booktitle={Conference on robot learning},
  pages={2226--2240},
  year={2023},
  organization={PMLR}
}

@inproceedings{khargonkar2023neuralgrasps,
  title={Neuralgrasps: Learning implicit representations for grasps of multiple robotic hands},
  author={Khargonkar, Ninad and Song, Neil and Xu, Zesheng and Prabhakaran, Balakrishnan and Xiang, Yu},
  booktitle={CoRL},
  pages={516--526},
  year={2023},
  organization={PMLR}
}

@inproceedings{cheng2023nod,
  title={NOD-TAMP: Multi-step manipulation planning with neural object descriptors},
  author={Cheng, Shuo and Garrett, Caelan Reed and Mandlekar, Ajay and Xu, Danfei},
  booktitle={CoRL 2023 Workshop on Learning Effective Abstractions for Planning (LEAP)},
  year={2023}
}

@article{kumar2021rma,
  title={RMA: Rapid motor adaptation for legged robots},
  author={Kumar, Ashish and Fu, Zipeng and Pathak, Deepak and Malik, Jitendra},
  journal={arXiv preprint arXiv:2107.04034},
  year={2021}
}

@inproceedings{zarrin2023hybrid,
  title={Hybrid learning-and model-based planning and control of in-hand manipulation},
  author={Zarrin, Rana Soltani and Jitosho, Rianna and Yamane, Katsu},
  booktitle={2023 IROS},
  pages={8720--8726},
  year={2023},
  organization={IEEE}
}

@article{jin2024complementarity,
  title={Complementarity-free multi-contact modeling and optimization for dexterous manipulation},
  author={Jin, Wanxin},
  journal={arXiv preprint arXiv:2408.07855},
  year={2024}
}

@inproceedings{cheng2021contact,
  title={Contact mode guided sampling-based planning for quasistatic dexterous manipulation in 2d},
  author={Cheng, Xianyi and Huang, Eric and Hou, Yifan and Mason, Matthew T},
  booktitle={ICRA},
  pages={6520--6526},
  year={2021},
  organization={IEEE}
}

@inproceedings{grady2021contactopt,
  title={Contactopt: Optimizing contact to improve grasps},
  author={Grady, Patrick and Tang, Chengcheng and Twigg, Christopher D and Vo, Minh and Brahmbhatt, Samarth and Kemp, Charles C},
  booktitle={CVPR},
  pages={1471--1481},
  year={2021}
}

@inproceedings{qi2017pointnet,
  title={Pointnet: Deep learning on point sets for 3d classification and segmentation},
  author={Qi, Charles R and Su, Hao and Mo, Kaichun and Guibas, Leonidas J},
  booktitle={CVPR},
  pages={652--660},
  year={2017}
}



\clearpage
\section*{APPENDIX}
\begin{figure*}
\vspace{0.2cm}
  \begin{subfigure}{0.235\textwidth}
    \centering
    \includegraphics[width=\linewidth]{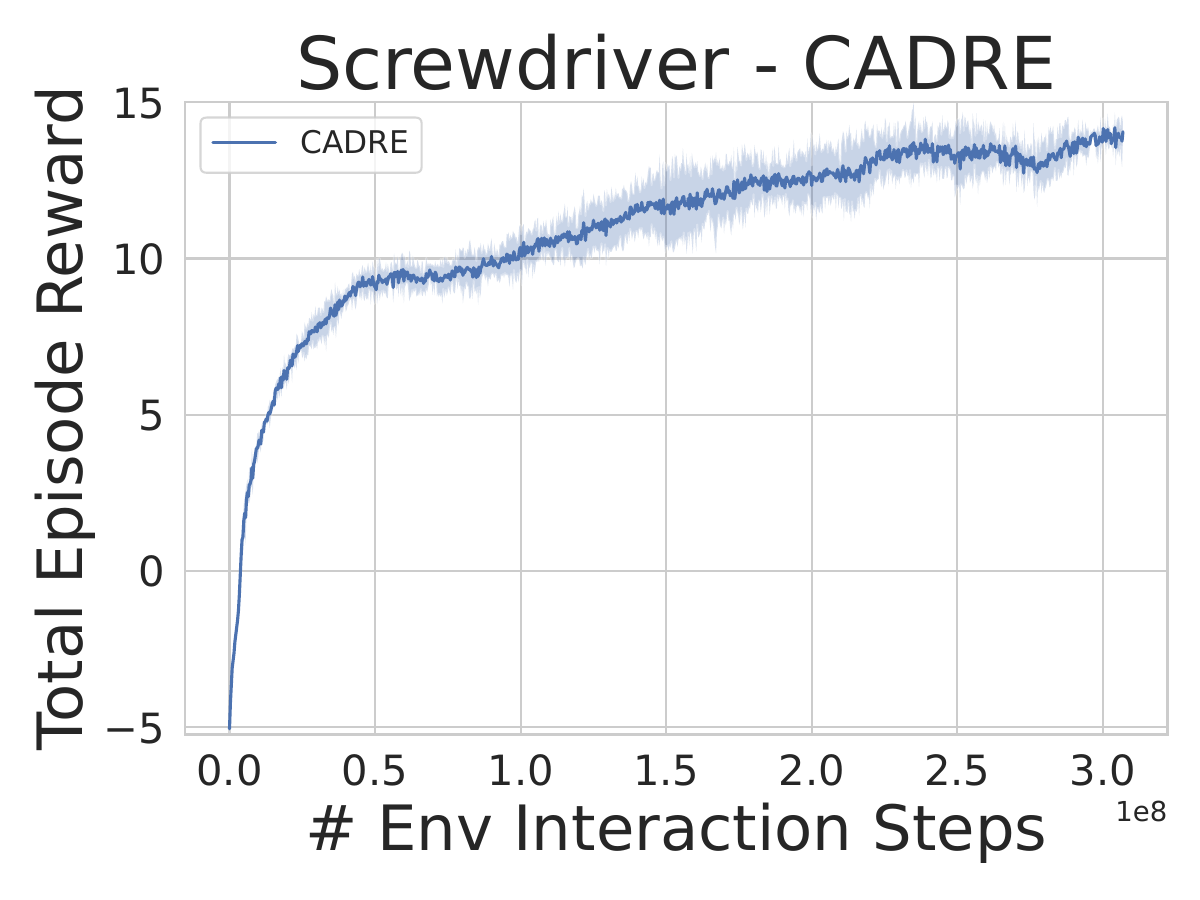}
  \end{subfigure}
  \hfill
  \begin{subfigure}{0.235\textwidth}
    \centering
    \includegraphics[width=\linewidth]{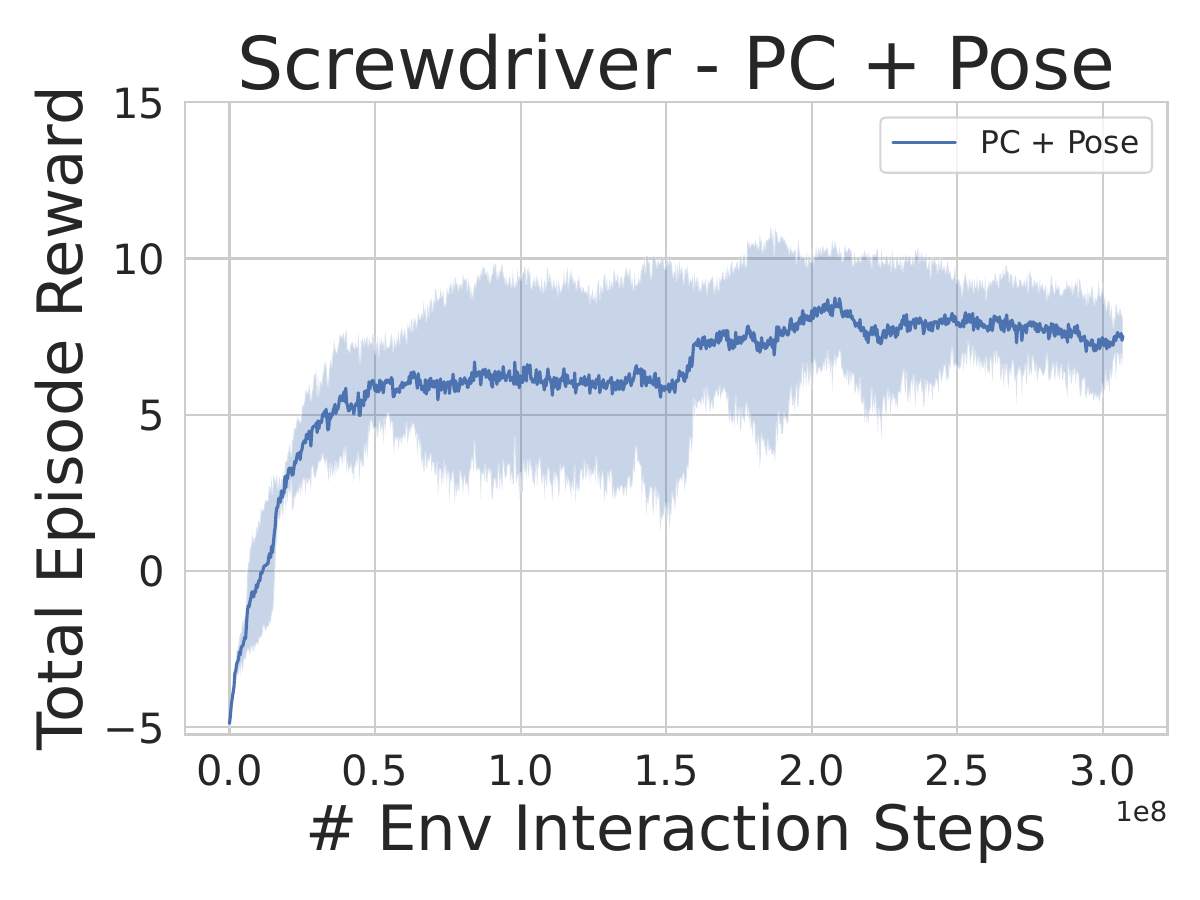}
  \end{subfigure}
  \hfill
  \begin{subfigure}{0.235\textwidth}
    \centering
    \includegraphics[width=\linewidth]{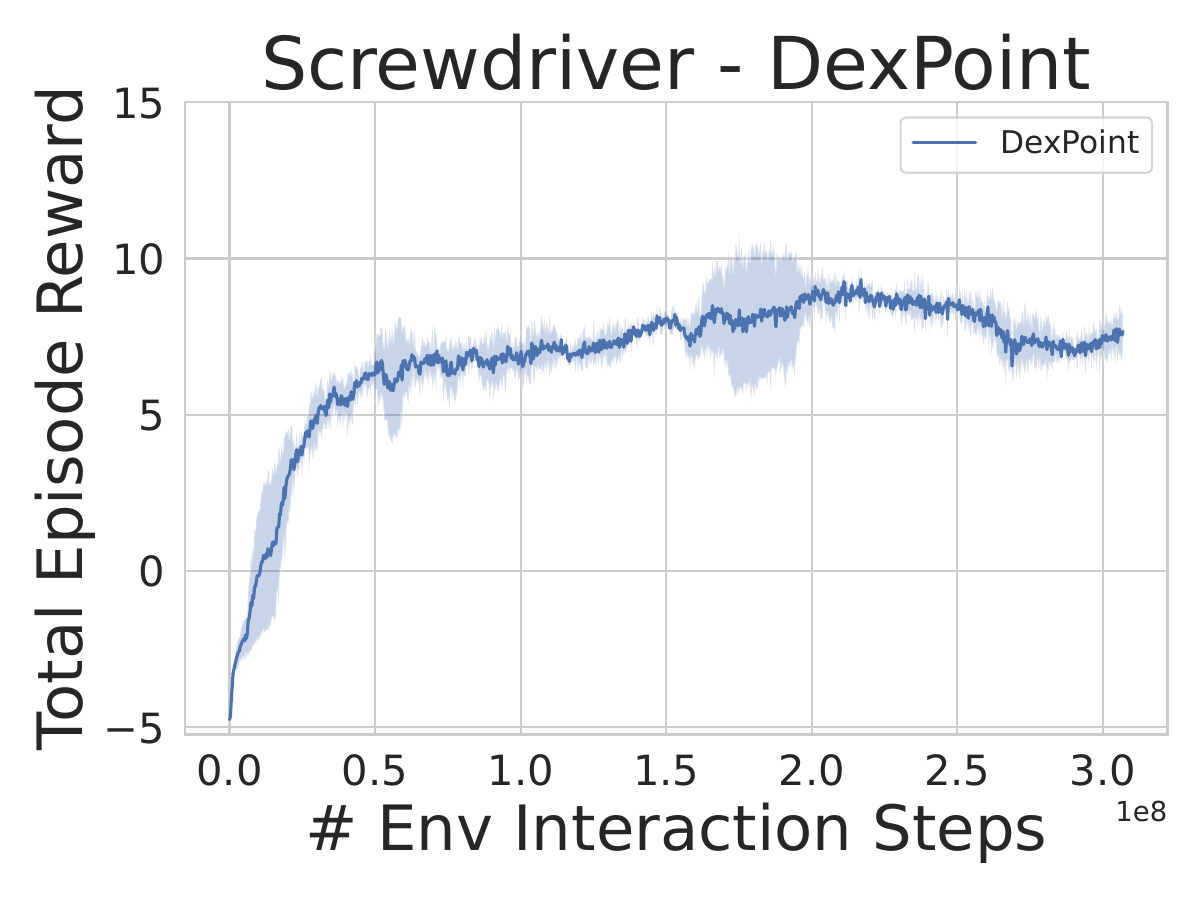}
  \end{subfigure}
  \hfill
  \begin{subfigure}{0.235\textwidth}
    \centering
    \includegraphics[width=\linewidth]{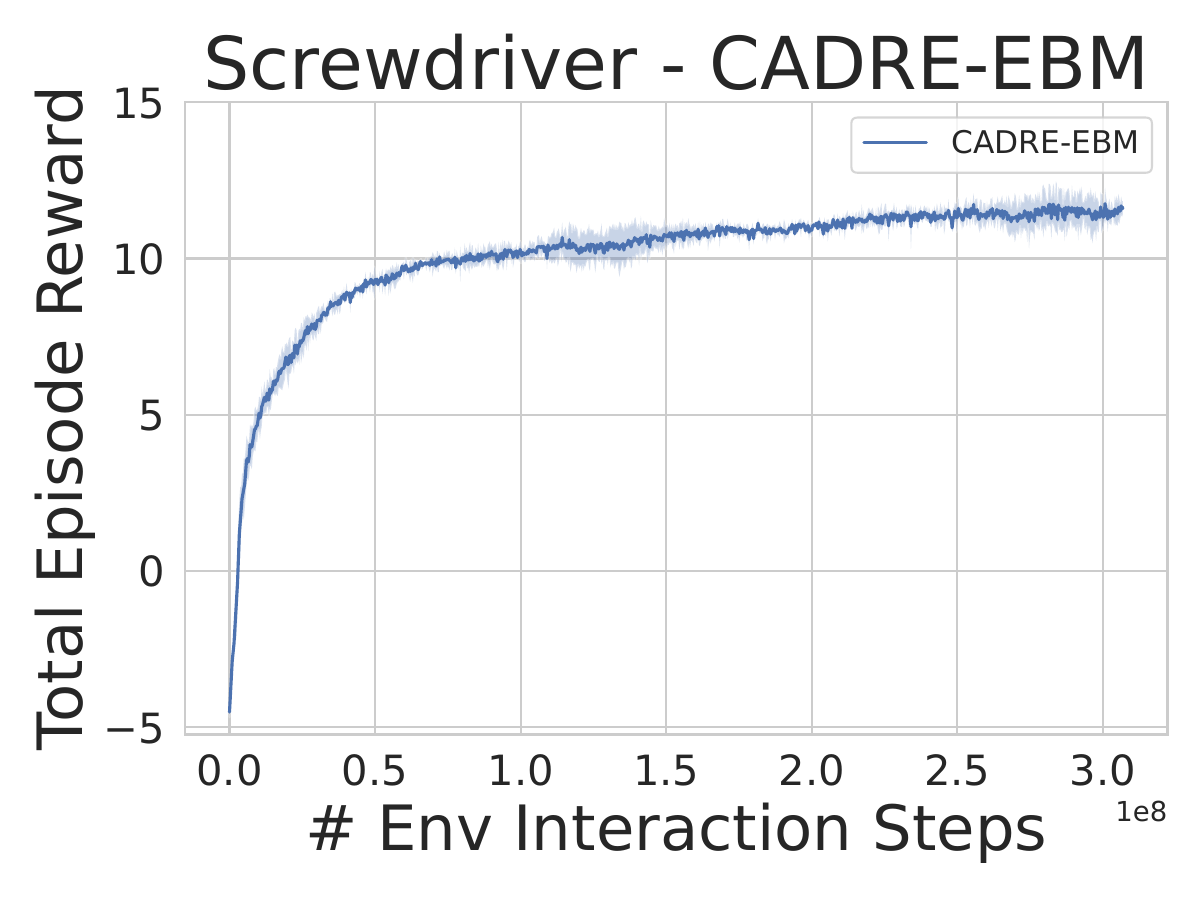}
  \end{subfigure}
  \hfill
  \begin{subfigure}{0.235\textwidth}
    \centering
    \includegraphics[width=\linewidth]{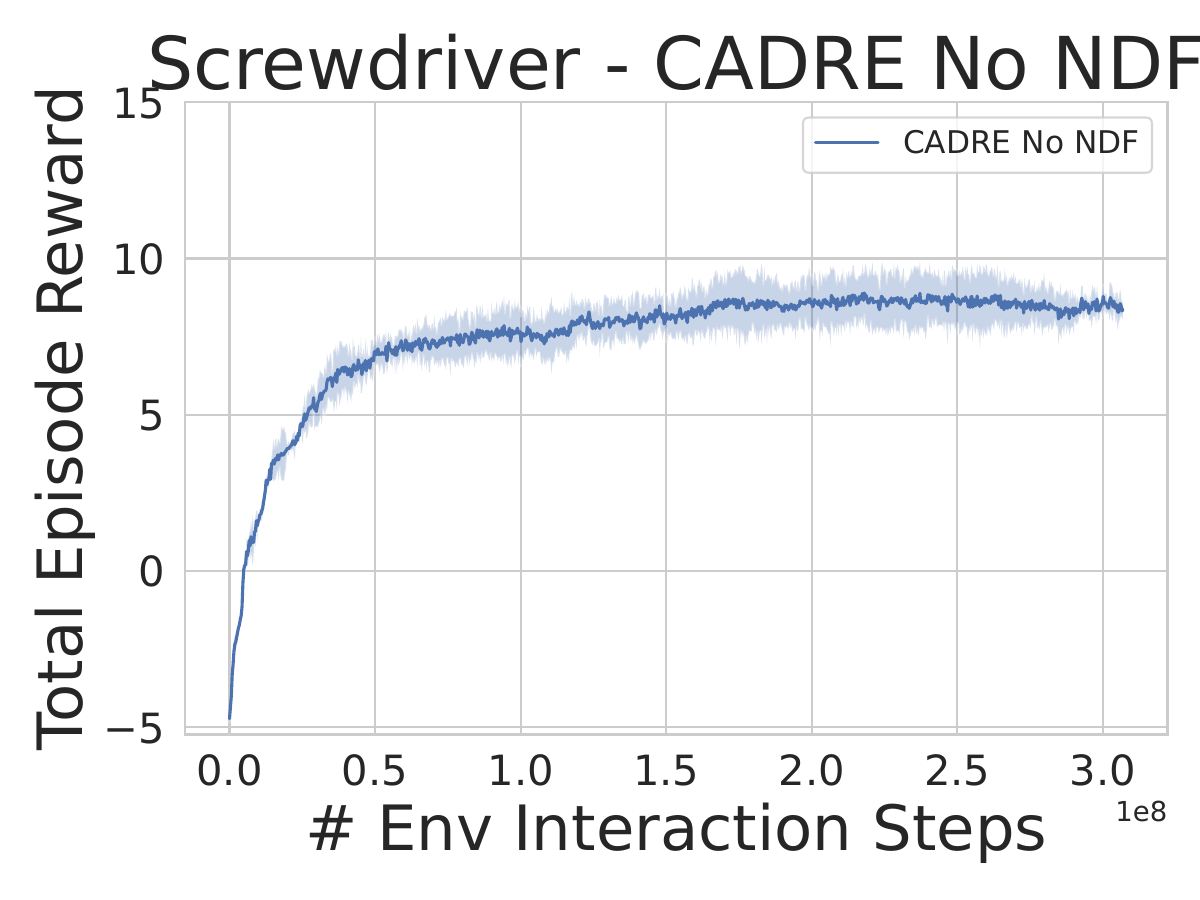}
  \end{subfigure}
  \hfill
  \begin{subfigure}{0.235\textwidth}
    \centering
    \includegraphics[width=\linewidth]{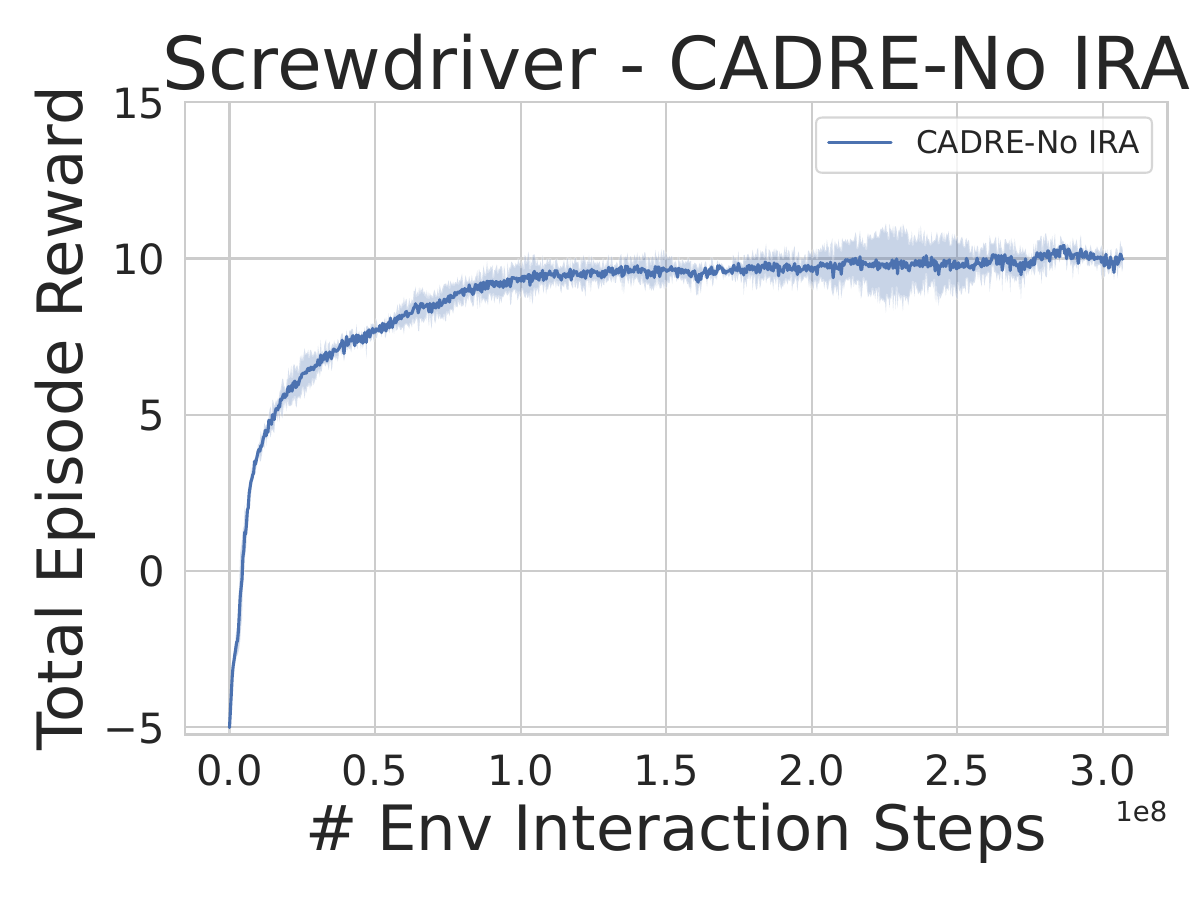}
  \end{subfigure}
    \begin{subfigure}{0.235\textwidth}
    \centering
    \includegraphics[width=\linewidth]{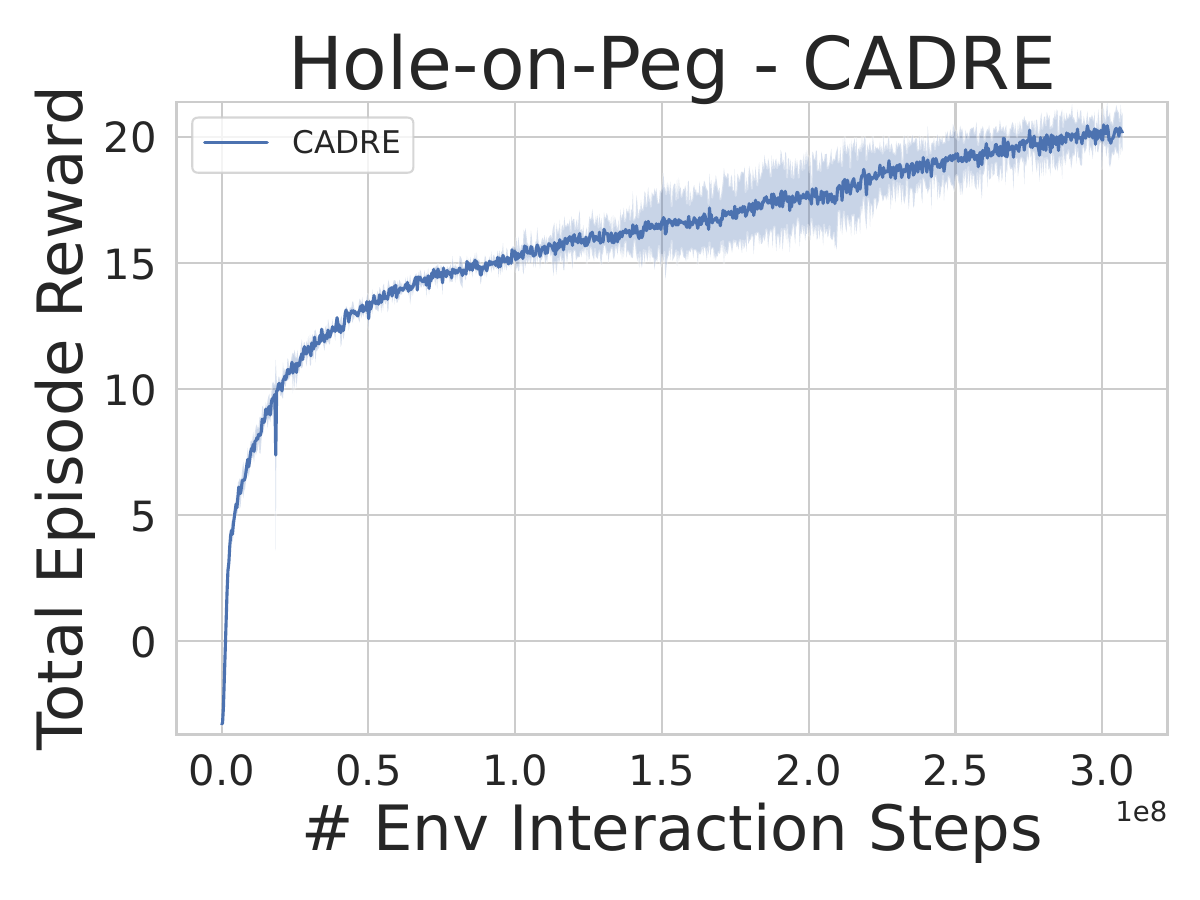}
  \end{subfigure}
  \hfill
  \begin{subfigure}{0.235\textwidth}
    \centering
    \includegraphics[width=\linewidth]{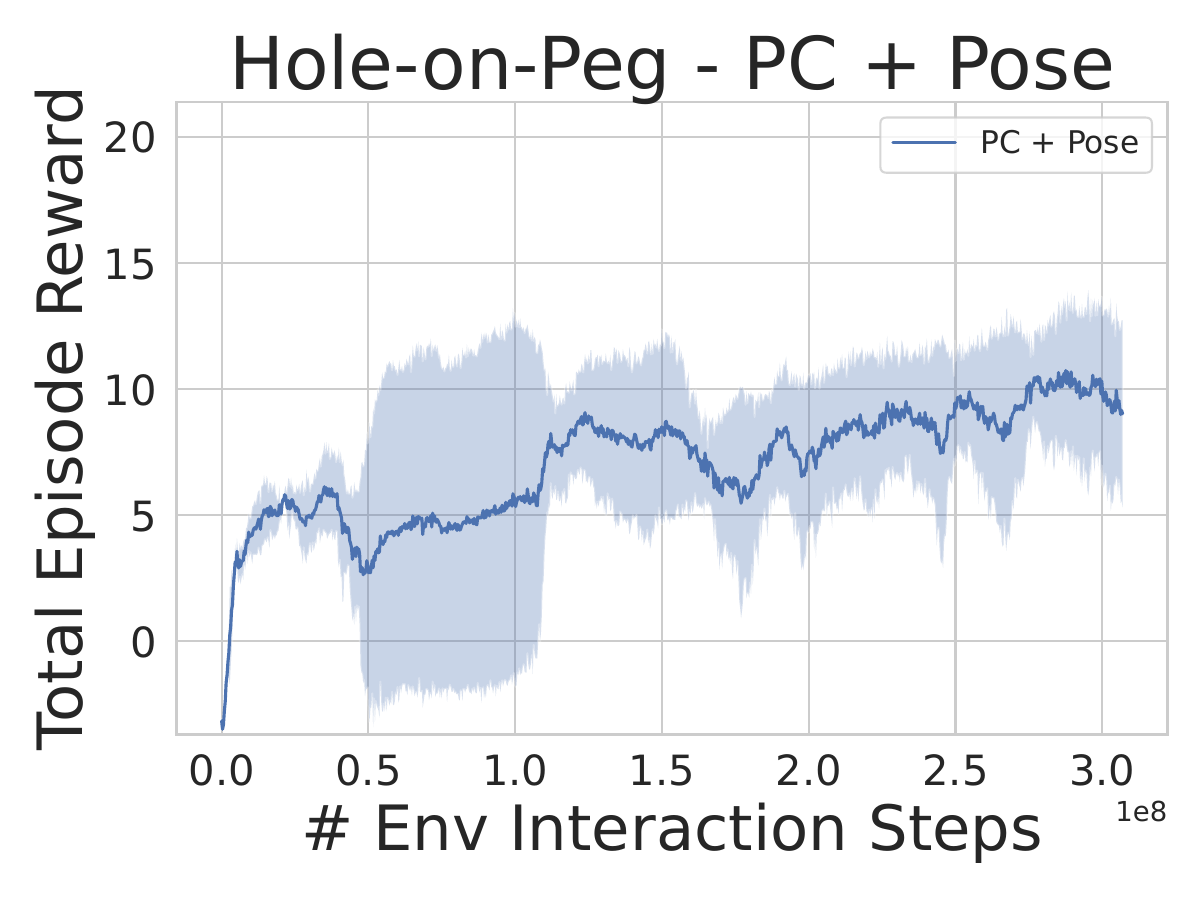}
  \end{subfigure}
  \hfill
  \begin{subfigure}{0.235\textwidth}
    \centering
    \includegraphics[width=\linewidth]{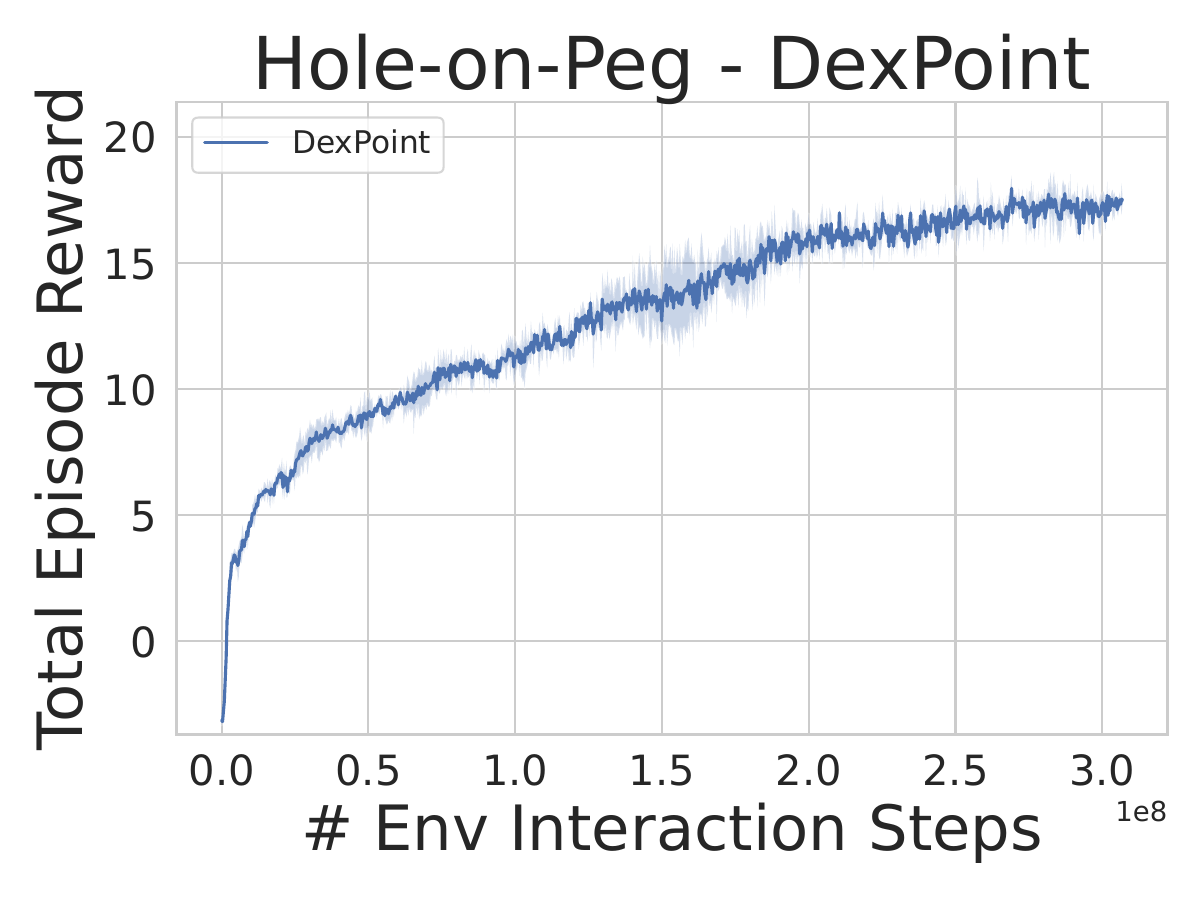}
  \end{subfigure}
  \hfill
  \begin{subfigure}{0.235\textwidth}
    \centering
    \includegraphics[width=\linewidth]{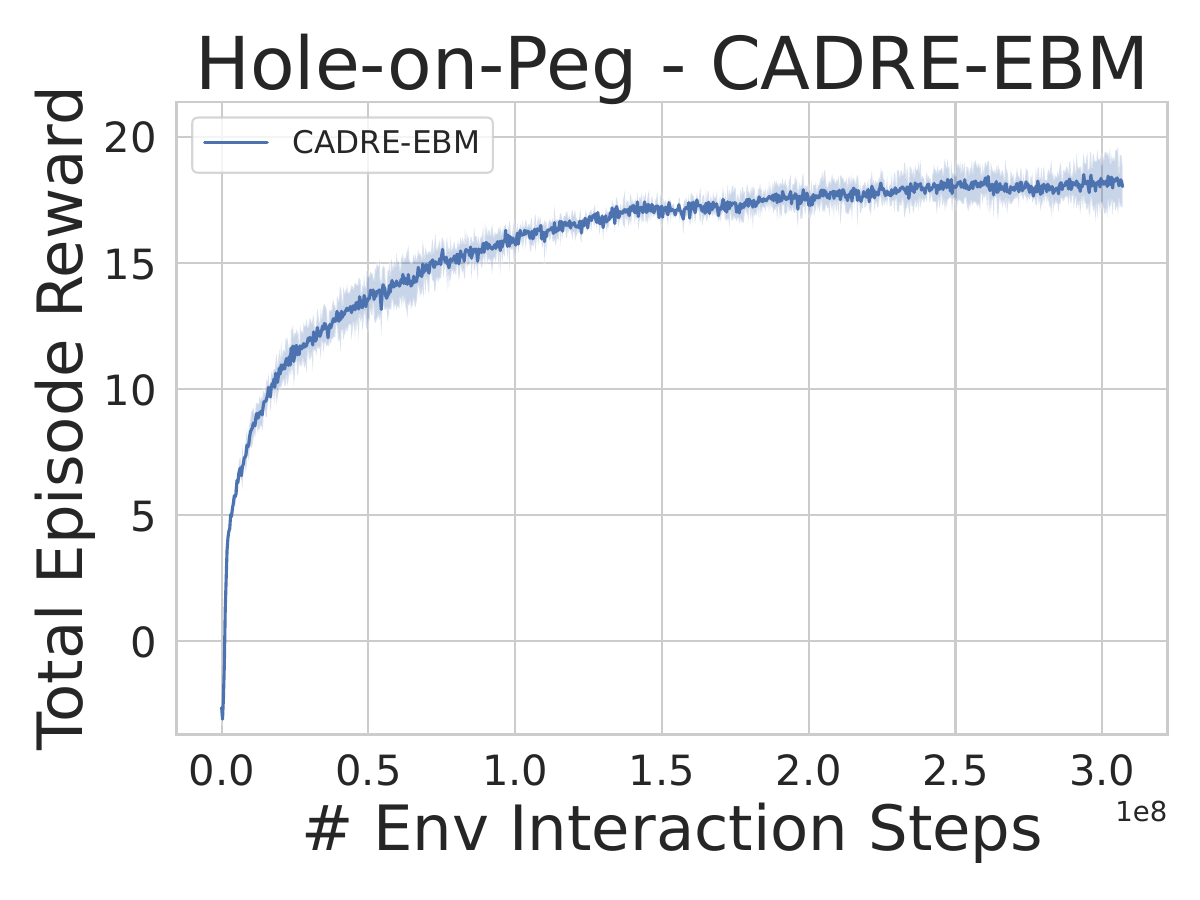}
  \end{subfigure}
  \hfill
  \begin{subfigure}{0.235\textwidth}
    \centering
    \includegraphics[width=\linewidth]{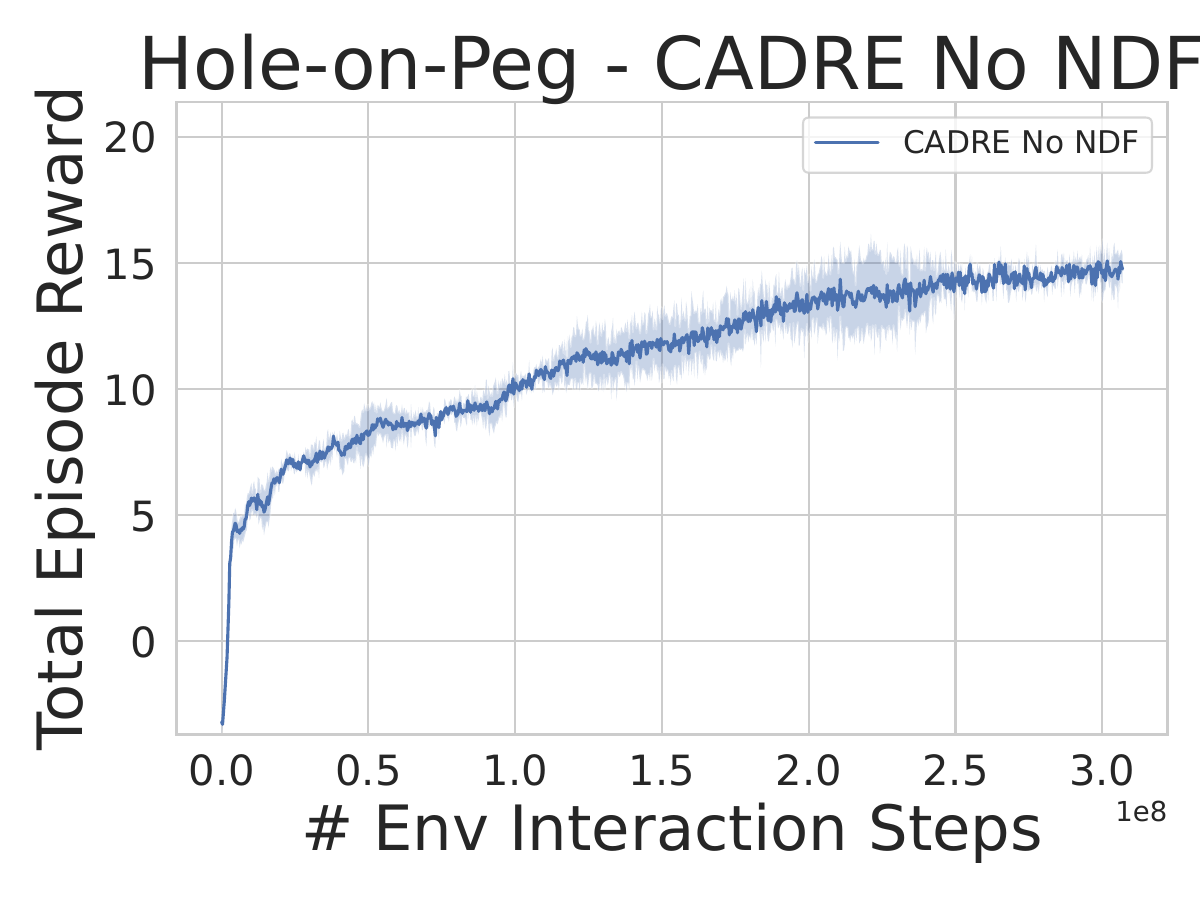}
  \end{subfigure}
  \hfill
  \begin{subfigure}{0.235\textwidth}
    \centering
    \includegraphics[width=\linewidth]{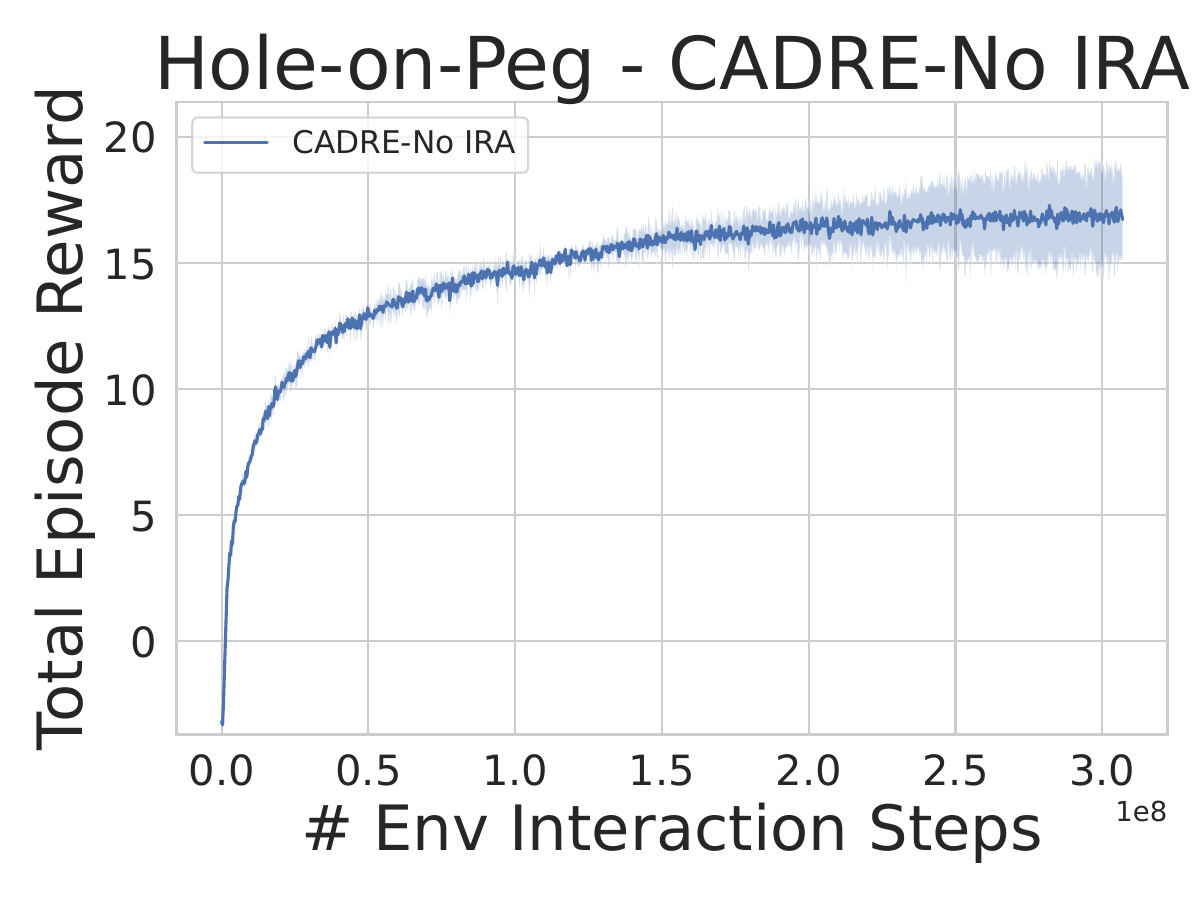}
  \end{subfigure}

  \caption{Three different seeds are used for RL training, and the average results are shown. The variance is relatively small as we use a large batch size during training. }
  \label{fig:curve}
\end{figure*}
\section{RL training curves}
The RL training curves are shown in Fig.~\ref{fig:curve}. CADRE achieves higher rewards than baselines given the same number of environment interactions. In the screwdriver recovery task, although the baseline methods have eventually converged, they fail to match CADRE's performance. For the hole-on-peg task, although all methods are nearing convergence, CADRE consistently outperforms the baselines.
\section{Domain Randomization} 
Since accurately perceiving fast-moving objects can be challenging in real-world scenarios, catching dynamic objects for recovery needs to handle significant uncertainties. To address this, we apply domain randomization~\cite{tobin2017domain} to improve robustness. 
Specifically, we introduce external disturbances and perception noise: at each time step, a random external wrench $\mathbf{w}_{ext}$ is applied at the center of the screwdriver, and random perception noise $\boldsymbol{\epsilon}$ sampled from a uniform distribution is added to the object's pose and velocities.

\section{Parameters for Generating Objects}
\label{app_sec:shape_param}
\begin{table*}
    \centering
    \begin{tabular}{|c|c|c|c|c|c|c|c|}
    \hline
           & \multicolumn{4}{c|}{Screwdriver} & \multicolumn{3}{c|}{Socket} \\
           \hline
           & handle r & handle l & shaft r & shaft l & inner r & outer r & thickness \\
           \hline
        ID & $\mathcal{U}(1, 3)$ & $\mathcal{U}(7, 14)$ & $\mathcal{U}(0.4, 1.4)$ & $\mathcal{U}(8, 10)$ & $\mathcal{U}(1, 2.5)$ & $\mathcal{U}(3, 4)$ & $\mathcal{U}(3, 5)$ \\
        \hline
        OOD & $\mathcal{U}(1.5, 3)$ & $\mathcal{U}(2, 4)$ & $\mathcal{U}(0.3, 0.7)$ & $\mathcal{U}(2, 4)$  & $\mathcal{U}(1, 2.5)$ & $\mathcal{U}(3, 4)$ & $\mathcal{U}(1, 3)$ \\
        \hline
    \end{tabular}
    \caption{Distribution used for generating object geometries. r: radius, l: length. $\mathcal{U}$ means uniform distribution. All units are in centimeters. OOD objects are generally shorter(screwdriver) or thinner(socket) than ID ones. }
    \label{tab:shape_distribution}
\end{table*}
For OOD object design, we specifically choose shorter screwdrivers and thinner sockets as those choices present significant challenges, where the desired contact regions are smaller and generally require more precise finger control. 




\end{document}